\documentclass[journal]{IEEEtran}
\usepackage{cite}
\usepackage{amsmath,amssymb,amsfonts}
\usepackage{graphicx}
\usepackage{array}
\usepackage{tabularx}
\usepackage{booktabs}
\usepackage{textcomp}
\usepackage{url}
\usepackage[caption=false,font=footnotesize]{subfig}
\usepackage{algorithmic}
\usepackage{dblfloatfix}
\usepackage{placeins}
\newcounter{algorithm}

\begin{document}

\title{GSRAIN: Physically Calibrated High-/Low-Frequency Rainfall Synthesis for 3D Gaussian Driving Scenes}
\author{Fanyu~Wang, Longgao~Zhang, and Junyi~Chen%
\thanks{Fanyu Wang, Longgao Zhang, and Junyi Chen are with the College of Automotive and Energy Engineering, Tongji University, Shanghai, China.}%
\thanks{Corresponding author: Junyi Chen.}%
}
\maketitle

\begin{abstract}
Existing rainfall simulation methods for autonomous driving remain limited in physical controllability and multi-view consistency. This paper presents GSRAIN, a high-/low-frequency rainfall synthesis method for 3D Gaussian Splatting (3DGS) driving scenes. GSRAIN constructs a high-frequency raindrop model from measured rainfall data and generates low-frequency rainy appearance using a geometry-aware single-step diffusion model. The two effects are then fused in a unified 3DGS scene, enabling rainfall-intensity control over the range of 0--13~mm/h. The proposed method achieves a Fr\'{e}chet Inception Distance (FID) of 149.09, outperforming CycleGAN-Turbo (155.71) and WeatherEdit (157.94). Object-detection and closed-loop driving experiments further show that the generated scenes expose scene-dependent performance changes of the evaluated algorithms under controllable rainfall. These results indicate that GSRAIN provides an effective approach for constructing physically controllable, repeatable, and closed-loop-compatible rainy-weather test scenes for autonomous driving.
\end{abstract}


\section{Introduction}
\label{sec:introduction}

\IEEEPARstart{A}{utonomous} driving has progressed rapidly with advances in artificial intelligence, sensing technologies, and vehicle--infrastructure cooperation \cite{ref1}. Nevertheless, the complexity and uncertainty of real-world road environments still require autonomous vehicles to undergo sufficient and systematic testing before deployment \cite{ref2,ref3}. System safety depends on reliable environmental perception and scene understanding. In both modular and end-to-end systems, perception errors may propagate to planning and control and eventually lead to unsafe behavior \cite{ref1,ref2}. Autonomous driving validation commonly combines virtual simulation, proving-ground tests, and public-road tests \cite{ref3}; however, physical testing remains constrained by safety risks, limited scenario coverage, high implementation cost, and poor repeatability, making it difficult to efficiently cover complex conditions and low-probability critical events \cite{ref2,ref3}.

Rainfall is a high-risk and frequently encountered trigger for vision-based autonomous driving. It reduces scene observability and changes object appearance, road-surface states, ambient illumination, and reflection patterns, thereby disturbing the recognition of and response to surrounding traffic conditions \cite{ref4}. Real rainfall is difficult to obtain and control on demand, while on-road testing is costly, time-consuming, and difficult to reproduce. Virtual simulation therefore provides a practical means of observing algorithm responses under controllable and repeatable rainfall conditions \cite{ref5,ref6}.

At a high level, virtual scene construction can be divided into graphics-driven and data-driven approaches. Graphics-driven methods offer explicit control but depend heavily on manually created assets. Data-driven methods include generative world models and scene reconstruction from real observations. The former can synthesize visually realistic scenes but are commonly designed for open-loop tasks, whereas the latter preserve real geometry and support novel-view rendering but provide limited controllable weather editing \cite{ref7}. Existing rainfall simulation methods also exhibit complementary limitations. Physics-based models support parameter control but cannot fully reproduce global rainy appearance, while learning-based transfer can generate rich appearance changes but may lack physical calibration and cross-view consistency \cite{ref8,ref9}. Consequently, it remains difficult to preserve real geometry, control numerical rainfall intensity, and generate multi-view-consistent rainy appearance within one unified 3D scene.

To address these issues, this paper proposes GSRAIN, a high-/low-frequency rainfall synthesis method based on 3DGS. The high-frequency branch fits raindrop diameter, velocity, and number density from measured rainfall data and generates near-field rain-streak Gaussians and far-field haze Gaussians under a target rainfall intensity. The low-frequency branch learns geometry-aware multi-view rainy appearance transfer and uses the transferred images to construct a low-frequency rainy 3DGS background. The two branches are independent, and their outputs are jointly rendered in a unified 3DGS scene through depth-aware alpha compositing. This design supports rainfall-intensity control over 0--13~mm/h and provides controllable and repeatable rainy scenes for autonomous driving evaluation.

The main contributions are summarized as follows:
\begin{itemize}
\item We propose GSRAIN, a high-/low-frequency decoupled rainfall synthesis framework for 3DGS driving scenes. Local raindrop effects and global rainy appearance are modeled separately and fused within a unified 3DGS representation and rendering process.
\item We construct a high-frequency raindrop model calibrated with measured rainfall data. The drop-size distribution, velocity--diameter relationship, and number density are explicitly mapped to the target rainfall intensity, and rain-streak and haze Gaussians enable graded rainfall synthesis over 0--13~mm/h.
\item We develop a low-frequency rainy-appearance transfer method that integrates surface-normal priors, cross-view feature matching, and mutual self-attention. The method improves multi-view appearance coordination while preserving the main scene structure, and the transferred results are reconstructed into the 3DGS scene.
\end{itemize}

\section{Related Work}
\label{sec:related-work}

\subsection{3DGS-Based Driving Scene Reconstruction}
\label{subsec:related-reconstruction}

Virtual driving scenes can be constructed using manually created assets and graphics engines or learned and reconstructed from real data. The former generates sensor observations through 3D models, materials, illumination, and physical rules. CARLA provides graphics rendering and sensor simulation, whereas SUMO and MetaDrive support traffic-flow and interactive-behavior modeling \cite{ref7,ref10,ref11}. These approaches facilitate explicit control of scene elements, but their modeling cost is high, and the reconstruction of real textures, materials, and complex environmental changes depends strongly on the quality of manually created assets.

To reduce manual modeling, data-driven methods learn scene distributions from real driving data and can be divided into generative world models and reconstruction from real observations. World models learn spatiotemporal evolution from large-scale datasets and generate images, videos, or other sensor observations \cite{ref12,ref13}; semantic maps, bird's-eye-view layouts, geographic information, and conditional prompts further improve generation controllability \cite{ref14,ref15}. However, current world models mainly target open-loop prediction or content generation and remain limited in explicit 3D representation and real-time environment updates driven by the system under test \cite{ref16}.

Scene-reconstruction methods directly recover 3D environments from real observations, preserving scene geometry and supporting novel-view rendering. NeRF and 3DGS are representative techniques \cite{ref17,ref18}. NeRF-based methods can synthesize camera and LiDAR observations, but frequent network queries limit rendering efficiency and generalization to large viewpoint changes \cite{ref19}. In contrast, 3DGS represents scenes with explicit Gaussian primitives and provides both high visual quality and real-time rendering \cite{ref18,ref20}. Recent studies have improved driving-scene reconstruction through dynamic-scene decomposition, geometric and semantic modeling, multi-sensor supervision, and completion of unobserved regions \cite{ref21,ref22,ref23,ref24}. Nevertheless, most existing methods focus on clear-weather reconstruction and do not fully support rainfall editing calibrated by measured data with numerical intensity control.

\subsection{Physics-Based Rain Rendering and Rainfall-Intensity Control}
\label{subsec:related-rain}

Rainfall effects can be divided into high-frequency local disturbances and low-frequency global appearance changes \cite{ref8,ref9,ref25}. High-frequency effects include rain streaks, local occlusion, and distance-dependent haze. Early 2D approaches synthesized rain streaks through post-processing or physical imaging models \cite{ref26,ref27}, while 3D approaches explicitly modeled raindrop particles or Gaussian primitives to improve geometric and temporal consistency. Halder \textit{et al.} generated rainfall particles according to raindrop appearance and motion \cite{ref28}; RainyGS and WeatherEdit used 3D Gaussians to represent rain streaks, splashes, or weather effects \cite{ref25,ref29}. However, measured drop-size spectra, velocity--diameter relationships, distance-dependent attenuation, and physically meaningful rainfall-intensity calibration have not yet been integrated into a unified driving-scene model.

Low-frequency rainfall appearance mainly includes wet road surfaces, enhanced specular reflections, reduced ambient brightness, and contrast changes. Two-dimensional image-translation methods learn mappings from clear to rainy domains using generative adversarial networks or diffusion models \cite{ref34,ref35,ref36,ref37}. They can produce rich rainy appearance, but independent processing of each view may alter scene content. Three-dimensional style-transfer methods optimize the appearance or geometry of NeRF or 3DGS while preserving content \cite{ref30,ref31,ref32,ref33}; StyleMe3D and FantasyStyle further improve multimodal control and semantic alignment \cite{ref31,ref32}, but their main objective is artistic stylization rather than physically grounded rainfall appearance.

For multi-view driving data, weather transfer must also coordinate texture, color, and local semantics across neighboring views; otherwise, subsequent 3DGS reconstruction may become unstable. Existing cross-view guidance can use feature matching or attention-based interaction to improve view consistency \cite{ref31,ref32,ref33,ref41}, but it has not yet been combined with measured rainfall-intensity calibration and real-time 3D rainfall rendering in a unified pipeline. GSRAIN therefore models high-frequency physical raindrops and low-frequency multi-view rainy appearance separately and fuses them in a 3DGS scene.

\section{Methodology}
\label{sec:methodology}

\subsection{Overview}
\label{subsec:overview}

\begin{figure*}[!t]
\centering
\includegraphics[width=0.98\textwidth]{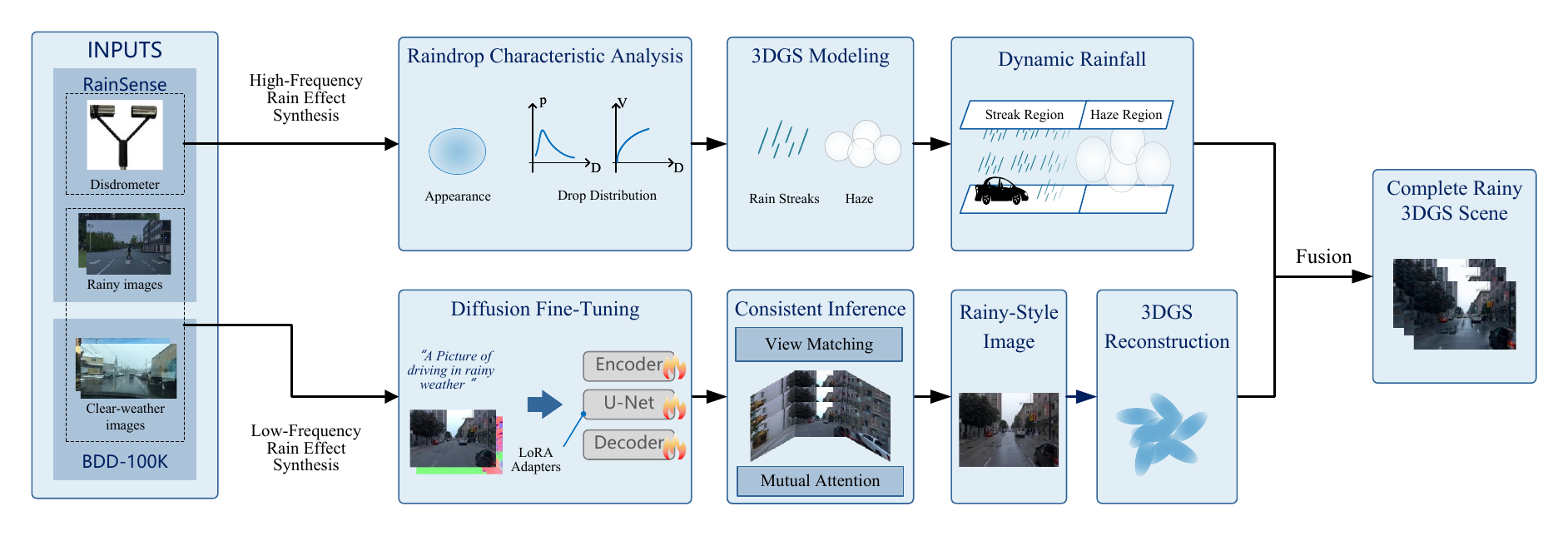}
\caption{Overview of the proposed GSRAIN framework. The high-frequency branch builds physically calibrated rain-streak and haze Gaussians, while the low-frequency branch produces geometry-aware multi-view rainy appearance and a rainy 3DGS background.}
\label{fig:framework}
\end{figure*}

As shown in Fig.~\ref{fig:framework}, GSRAIN contains two independent branches for low-frequency rainy appearance and high-frequency raindrop effects. The low-frequency branch learns geometry-constrained multi-view rainy appearance transfer and reconstructs or re-optimizes a rainy 3DGS background from the transferred images. The high-frequency branch fits raindrop diameter, velocity, and number density from measured rainfall data and generates near-field rain-streak Gaussians and far-field haze Gaussians under a target rainfall intensity. During rendering, the rain-streak, haze, and background Gaussians jointly participate in depth-ordered alpha compositing, producing a complete rainy 3DGS scene with both global rainy appearance and local raindrop disturbances.

\subsection{Preliminaries on 3DGS}
\label{subsec:3dgs}

Three-dimensional Gaussian Splatting represents a scene with a large set of 3D ellipsoids. In 3D Euclidean space, a Gaussian distribution is defined by a mean vector and a symmetric positive-definite covariance matrix:
\begin{equation}
G(\mathbf{x})=\frac{1}{\sqrt{(2\pi)^3|\boldsymbol{\Sigma}|}}
\exp\!\left[-\frac{1}{2}(\mathbf{x}-\boldsymbol{\mu})^{\mathsf T}
\boldsymbol{\Sigma}^{-1}(\mathbf{x}-\boldsymbol{\mu})\right],
\label{eq:gaussian}
\end{equation}
where $\boldsymbol{\mu}$ is the Gaussian center and $\boldsymbol{\Sigma}$ controls its spatial extent and anisotropy.

Each Gaussian ellipsoid contains learnable attributes describing its geometry, appearance, and opacity contribution. The $i$th primitive is written as
\begin{equation}
\mathcal{G}_i=\{\boldsymbol{\mu}_i,\boldsymbol{\Sigma}_i,\mathbf{c}_i,\alpha_i\},
\label{eq:primitive}
\end{equation}
where $\mathbf{c}_i$ and $\alpha_i$ denote color (or view-dependent color parameters) and opacity, respectively.

\subsection{High-Frequency Raindrop Imaging with Gaussian Ellipsoids}
\label{subsec:high-frequency}

This section models high-frequency rainfall effects using 3D Gaussian ellipsoids. We first analyze raindrop appearance and statistical distribution, then construct near-field rain-streak and far-field haze models, and finally implement a dynamic rainfall process in 3D space.

\subsubsection{Raindrop Characteristics}
\label{subsubsec:raindrop-characteristics}

\paragraph{Appearance Characteristics}
The visual appearance of a raindrop is jointly determined by its color, scale, opacity, and distance-dependent attenuation. Because a raindrop is transparent, its observed color mainly comes from refracted environmental light \cite{ref8}. We approximate the raindrop color as
\begin{equation}
\mathbf{c}=\frac{1}{N_{\mathrm{sky}}}\sum_{p\in\Omega_{\mathrm{sky}}}\mathbf{I}(p),
\label{eq:color}
\end{equation}
where $\Omega_{\mathrm{sky}}$ is the sky region and $N_{\mathrm{sky}}$ is its number of pixels.

For most visual-recognition tasks, a falling raindrop can be assumed to reach an equilibrium shape. Experimental studies show that its equilibrium axis ratio varies with the original diameter $D$ \cite{ref26}:
\begin{equation}
\tau(D)=\frac{z}{x}=kD+b,\qquad 1~\mathrm{mm}<D<9~\mathrm{mm},
\label{eq:axis-ratio}
\end{equation}
where $z$ and $x$ are the maximum widths parallel and perpendicular to the velocity direction, respectively. The fitting coefficients are approximately $k=-0.07$ and $b=1.07$.

The opacity of a raindrop on the camera image plane depends on its physical size, falling velocity, and the exposure time \cite{ref27}:
\begin{equation}
\alpha_0=\frac{D}{V T_{\mathrm{exp}}},
\label{eq:opacity}
\end{equation}
where $T_{\mathrm{exp}}$ denotes the camera exposure time.

\begin{figure}[!t]
\centering
\includegraphics[width=0.96\linewidth]{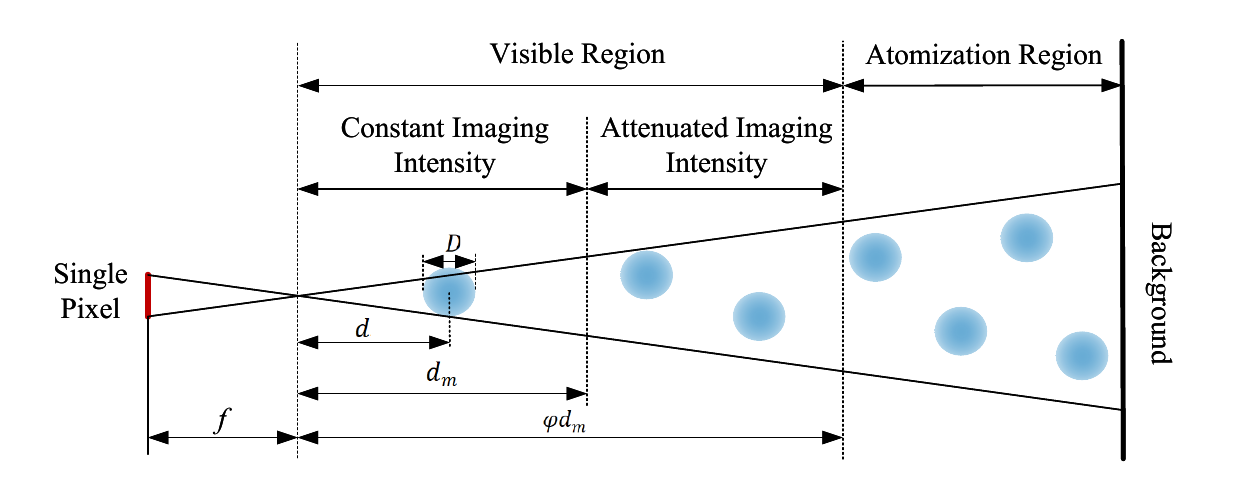}
\caption{Raindrop imaging regions and distance-dependent attenuation.}
\label{fig:raindrop-imaging}
\end{figure}

Because of the finite pixel resolution, only raindrops in a visible range form identifiable rain streaks. More distant droplets gradually become haze-like disturbances, as illustrated in Fig.~\ref{fig:raindrop-imaging}. Given the diameter $D$ and focal length $f$ in pixels, the distance at which the droplet exactly covers one pixel is
\begin{equation}
d_m=fD.
\label{eq:visible-distance}
\end{equation}
We model the distance-dependent opacity using
\begin{equation}
\alpha(d)=
\begin{cases}
\alpha_0, & d\le d_m,\\
\dfrac{d_m}{d}\alpha_0, & d_m<d\le \varphi d_m,\\
\alpha_m, & d>\varphi d_m,
\end{cases}
\label{eq:attenuation}
\end{equation}
where $\varphi$ is a critical haze coefficient determined by scene brightness and camera sensitivity, and $\alpha_m$ is the opacity assigned to the haze region.

\paragraph{Distribution Characteristics}
Under a target rainfall intensity $R$, the raindrop-diameter distribution is non-uniform. A gamma distribution is used to fit the measured drop-size spectrum \cite{ref39}:
\begin{equation}
p_D(D\mid R)=
\frac{D^{\mu_R-1}}{\Gamma(\mu_R)\theta_R^{\mu_R}}
\exp\!\left(-\frac{D}{\theta_R}\right),\qquad D>0,
\label{eq:gamma}
\end{equation}
where $D$ is the raindrop diameter, $p_D(D\mid R)$ is its probability density under target rainfall intensity $R$, $\mu_R$ is the dimensionless shape parameter, and $\theta_R$ is the scale parameter in millimeters.

Raindrop velocity is positively correlated with diameter. A quadratic function fitted to the mean velocities measured by RainSense is
\begin{equation}
V_{\mu}(D)=-0.342D^2+3.379D+0.340,
\label{eq:velocity}
\end{equation}
with a coefficient of determination of $R^2=0.997$.

\subsubsection{Raindrop Imaging Models}
\label{subsubsec:imaging-models}

\paragraph{Rain-Streak Model}
A moving rain streak can be viewed as a sequence of local highlights accumulated along the motion direction. Given velocity $V$ and exposure time $T_{\mathrm{exp}}$, the streak length is
\begin{equation}
L=VT_{\mathrm{exp}}.
\label{eq:streak-length}
\end{equation}
To limit the number of Gaussian ellipsoids used for each streak, the axial scale and duplication number are set as
\begin{equation}
s_z=\frac{L}{n},
\label{eq:streak-scale}
\end{equation}
\begin{equation}
n=\min\!\left(\frac{L}{\tau D},n_{\max}\right),
\label{eq:duplication}
\end{equation}
where $n_{\max}$ is the maximum duplication number.

\paragraph{Haze Model}
In the far field, individual raindrops are no longer visually distinguishable and instead appear as a continuous semi-transparent medium. GSRAIN therefore uses a low-density set of large, isotropic Gaussian ellipsoids with low opacity. This strategy reproduces spatially non-uniform haze while keeping the computational cost moderate.

\subsubsection{Dynamic Rainfall Simulation}
\label{subsubsec:dynamic-rain}

To simulate continuously falling raindrops, a dynamic rainfall process is constructed from the statistical distribution and Gaussian appearance models. The process consists of raindrop initialization, position updating, and boundary resetting. Repeated execution of these steps produces temporally stable and continuous rainfall that follows the prescribed statistics. Algorithm~\ref{alg:dynamic-rain} summarizes the complete process.

\begin{figure}[!t]
\centering
\begin{minipage}{0.98\columnwidth}
\refstepcounter{algorithm}\label{alg:dynamic-rain}
\noindent\footnotesize\textbf{Algorithm~\thealgorithm.} Dynamic rainfall simulation based on 3DGS.
\vspace{2pt}
\hrule
\vspace{2pt}
\footnotesize
\begin{algorithmic}[1]
\STATE \textbf{Stage 1: Rainfall-state initialization}
\STATE Determine $\bar{N}(R)$ and the gamma parameters $\mu_R,\theta_R$ from the target rainfall intensity $R$
\STATE Compute $V_k,V_f$ and the primitive counts $N=\bar{N}(R)V_k$ and $M$
\STATE Initialize $\mathcal{G}^{\mathrm{streak}}\leftarrow\varnothing$ and $\mathcal{G}^{\mathrm{fog}}\leftarrow\varnothing$
\FOR{$i=1$ to $N$}
  \STATE Sample $\mathbf{p}_i\in\mathcal{S}_k$ and diameter $D_i$; compute $\mathbf{v}_i$ and the remaining physical and appearance parameters
  \STATE $\mathcal{G}^{\mathrm{streak}}\leftarrow\mathcal{G}^{\mathrm{streak}}\cup\{(\mathbf{p}_i,D_i,\mathbf{v}_i,\mathbf{s}_i,\alpha_i,\mathbf{c}_i)\}$
\ENDFOR
\FOR{$j=1$ to $M$}
  \STATE Sample $\mathbf{p}_j\in\mathcal{S}_f$; assign fixed scale, opacity, and color, with $\mathbf{v}_j=\mathbf{0}$
  \STATE $\mathcal{G}^{\mathrm{fog}}\leftarrow\mathcal{G}^{\mathrm{fog}}\cup\{(\mathbf{p}_j,\mathbf{s}_j,\alpha_j,\mathbf{c}_j)\}$
\ENDFOR
\STATE \textbf{Stage 2: Temporal update and boundary reset}
\FOR{$t=1$ to $T$}
  \FOR{each $\mathcal{G}^{\mathrm{streak}}_i\in\mathcal{G}^{\mathrm{streak}}$}
    \STATE $\mathbf{p}_i\leftarrow\mathbf{p}_i+\mathbf{v}_i\Delta t$
    \IF{$\mathbf{p}_i\notin\mathcal{S}_k$}
      \STATE Resample $\mathbf{p}_i$ in the upper reset region $\widehat{\mathcal{S}}_k$
    \ENDIF
  \ENDFOR
  \STATE Randomly reset fog-Gaussian positions to model non-uniform perturbations
  \STATE $\mathcal{G}^{\mathrm{rain}}(t)\leftarrow\mathcal{G}^{\mathrm{streak}}\cup\mathcal{G}^{\mathrm{fog}}$
  \STATE Return $\mathcal{G}^{\mathrm{rain}}(t)$
\ENDFOR
\end{algorithmic}
\vspace{2pt}
\hrule
\end{minipage}
\end{figure}

\subsection{Low-Frequency Global Style Transfer with a Single-Step Diffusion Model}
\label{subsec:low-frequency}

The high-frequency branch explicitly models raindrop-induced effects in 3DGS. The low-frequency branch performs realistic and cross-view-consistent rainy style transfer in 2D, and the resulting rainy images supervise re-optimization of the original 3DGS scene so that global rainy appearance is injected into the 3D representation.

\subsubsection{Fine-Tuning the Single-Step Diffusion Model}
\label{subsubsec:fine-tuning}

\begin{figure*}[!t]
\centering
\includegraphics[width=0.88\textwidth]{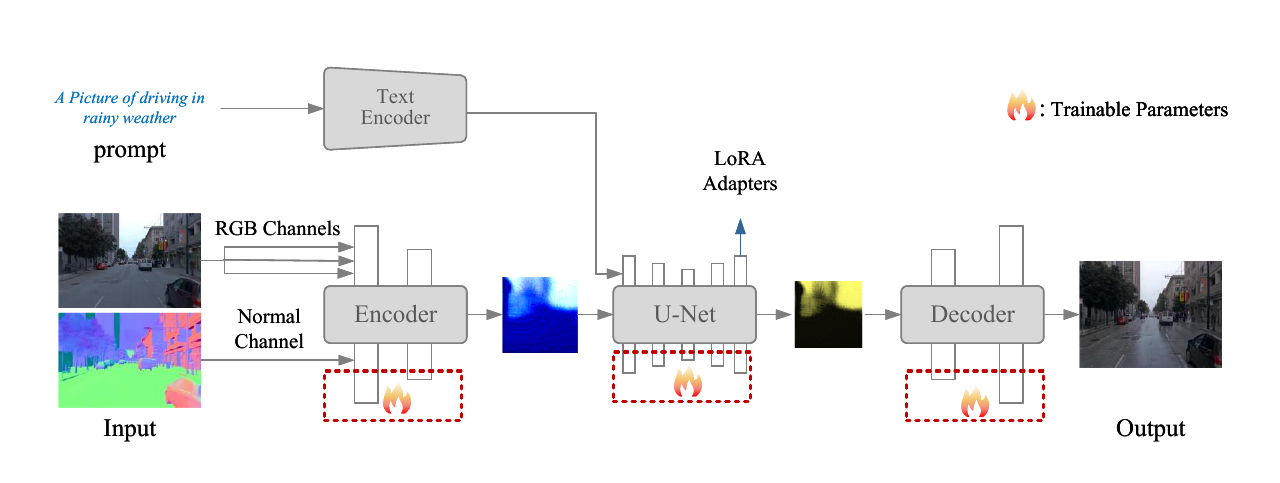}
\caption{Fine-tuning pipeline of the normal-conditioned single-step diffusion model.}
\label{fig:diffusion-finetuning}
\end{figure*}

We fine-tune CycleGAN-Turbo \cite{ref37} using LoRA \cite{ref38} to translate clear traffic scenes into rainy scenes. As shown in Fig.~\ref{fig:diffusion-finetuning}, a normal-map channel is added to the three RGB channels, forming a four-channel input to the VAE encoder. This geometry-aware input provides additional surface-orientation priors and improves the modeling of low-frequency rainy appearance.

Surface normals are estimated using DSINE \cite{ref40}. DSINE first predicts a pixel-level normal distribution, after which the 3D normal representation is converted into a single-channel grayscale map and fed to the VAE encoder as the fourth channel. The encoder can therefore use both appearance and local surface orientation to build a more stable representation of scene geometry.

\subsubsection{Consistency Enhancement for Style Transfer}
\label{subsubsec:consistency}

A diffusion model usually processes each frame independently, focusing on content preservation and style injection within the current image. When directly applied to multi-view driving scenes, it may produce inconsistent textures, colors, and local details across views. GSRAIN therefore combines cross-view feature matching with mutual self-attention. The former establishes geometric correspondences, while the latter enables information exchange among aligned view features inside the denoising network.

\paragraph{Cross-View Feature Matching}
Let neighboring view $a$ and current view $c$ have intrinsic matrix $\mathbf{K}$ and extrinsic parameters $(\mathbf{R}_a,\mathbf{t}_a)$ and $(\mathbf{R}_c,\mathbf{t}_c)$. For pixel $\widetilde{\mathbf{p}}_a$ with depth $d_a$, the corresponding 3D point in the neighboring camera coordinates is
\begin{equation}
\mathbf{X}_a=d_a\mathbf{K}^{-1}\widetilde{\mathbf{p}}_a.
\label{eq:backproject}
\end{equation}
It is transformed into world coordinates by
\begin{equation}
\mathbf{X}_w=\mathbf{R}_a^{-1}(\mathbf{X}_a-\mathbf{t}_a),
\label{eq:world}
\end{equation}
and reprojected to the current image plane as
\begin{equation}
\widetilde{\mathbf{p}}_c\sim\mathbf{K}(\mathbf{R}_c\mathbf{X}_w+\mathbf{t}_c).
\label{eq:reproject}
\end{equation}
Applying this transformation to valid pixels yields a reference feature map aligned with the current view.

\paragraph{Mutual Self-Attention}
Following \cite{ref41}, the current-view feature $\mathbf{F}_{v_c}$ is mapped to queries,
\begin{equation}
\mathbf{Q}_{v_c}=\mathbf{W}_Q\mathbf{F}_{v_c},
\label{eq:query}
\end{equation}
while the current feature and geometrically aligned neighboring feature $\mathbf{F}_{v_{a\rightarrow c}}$ are concatenated and mapped to keys and values:
\begin{equation}
\mathbf{K}_m=\mathbf{W}_K[\mathbf{F}_{v_{a\rightarrow c}};\mathbf{F}_{v_c}],
\label{eq:key}
\end{equation}
\begin{equation}
\mathbf{V}_m=\mathbf{W}_V[\mathbf{F}_{v_{a\rightarrow c}};\mathbf{F}_{v_c}].
\label{eq:value}
\end{equation}
The mutual attention is
\begin{equation}
\operatorname{Attn}_{\mathrm{mutual}}=
\operatorname{softmax}\!\left(\frac{\mathbf{Q}_{v_c}\mathbf{K}_m^{\mathsf T}}{\sqrt{d}}\right)\mathbf{V}_m,
\label{eq:attention}
\end{equation}
where $d$ is the feature-channel dimension.

\begin{figure*}[!t]
\centering
\includegraphics[width=0.88\textwidth]{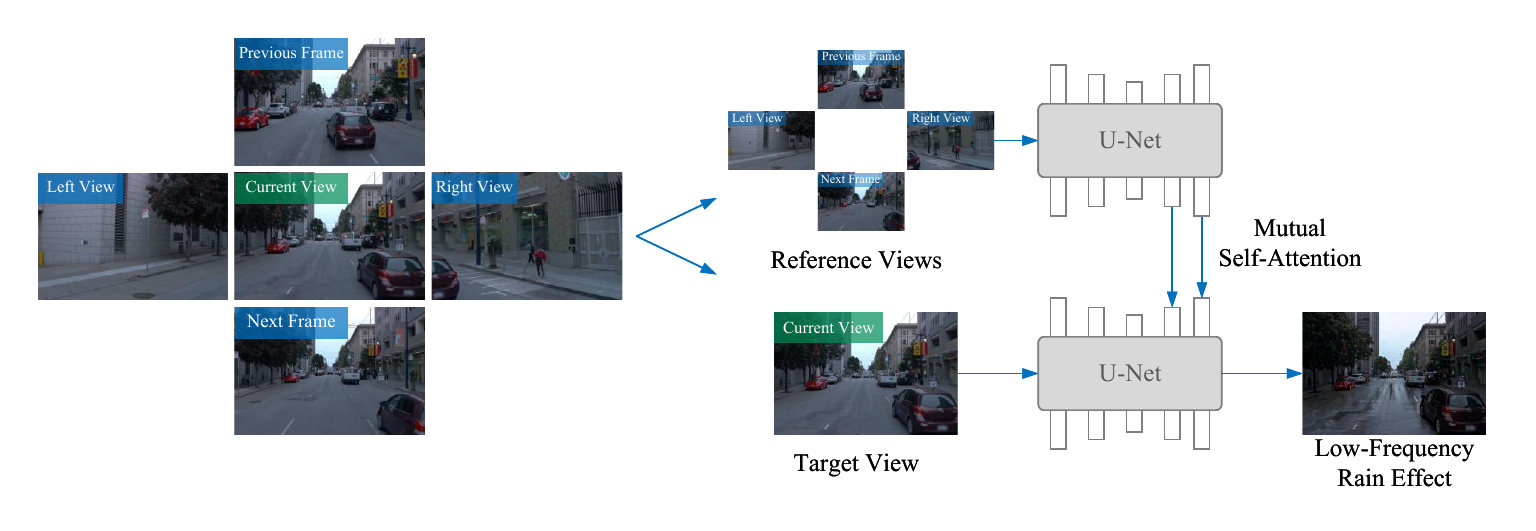}
\caption{Cross-view consistency enhancement using feature matching and mutual self-attention.}
\label{fig:mutual-attention}
\end{figure*}

\section{Experiments}
\label{sec:experiments}

This section introduces the datasets, training settings, and simulation configurations, and then evaluates low-frequency rainfall transfer, overall rainfall synthesis, and downstream autonomous-driving applications.

\subsection{Low-Frequency Rainfall Transfer}
\label{subsec:low-frequency-evaluation}

\subsubsection{Experimental Setup}
\label{subsubsec:low-frequency-setup}

RainSense and BDD-100K \cite{ref42} are used to construct the style-transfer dataset. The training and test splits account for 80\% and 20\%, respectively, as summarized in Table~\ref{tab:dataset}.

\begin{table}[!t]
\caption{Dataset Composition}
\label{tab:dataset}
\centering
\footnotesize
\begin{tabular}{lccc}
\toprule
Dataset & Clear & Rainy & Total \\
\midrule
RainSense & 22 & 22 & 44 \\
BDD-100K \cite{ref42} & 139 & 139 & 278 \\
Total & 161 & 161 & 322 \\
\bottomrule
\end{tabular}
\end{table}

The single-step diffusion model is fine-tuned for 20,000 iterations on one NVIDIA H800 GPU with 80~GB memory, taking approximately 16 hours. Both RGB images and normal maps are resized to $512\times512$ pixels, and the batch size is 1. The loss weights $\lambda_{\mathrm{cycle}}$, $\lambda_{\mathrm{idt}}$, and $\lambda_{\mathrm{GAN}}$ are set to 1.0, 1.0, and 0.5, respectively. CycleGAN-Turbo \cite{ref37}, Gemini 3.1 Flash Image \cite{ref43}, and WeatherEdit \cite{ref25} are used for comparison.

\subsubsection{Qualitative Comparison}
\label{subsubsec:qualitative}

\begin{figure*}[!t]
\centering
\setlength{\tabcolsep}{1.4pt}
\renewcommand{\arraystretch}{0.98}
\begin{tabular}{@{}>{\raggedleft\arraybackslash}m{0.12\textwidth}@{\hspace{3pt}}ccc@{}}
\footnotesize (a) Clear input &
\includegraphics[width=0.220\textwidth]{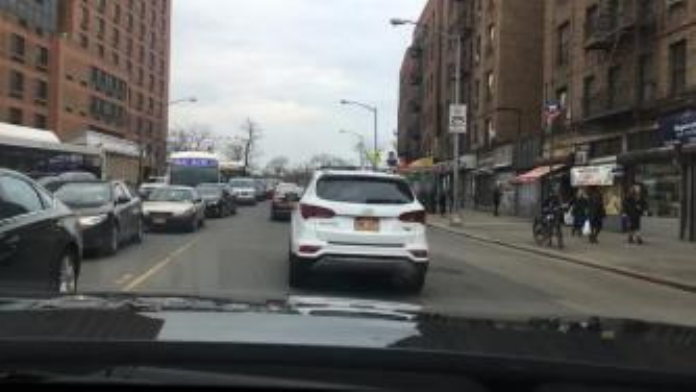} &
\includegraphics[width=0.220\textwidth]{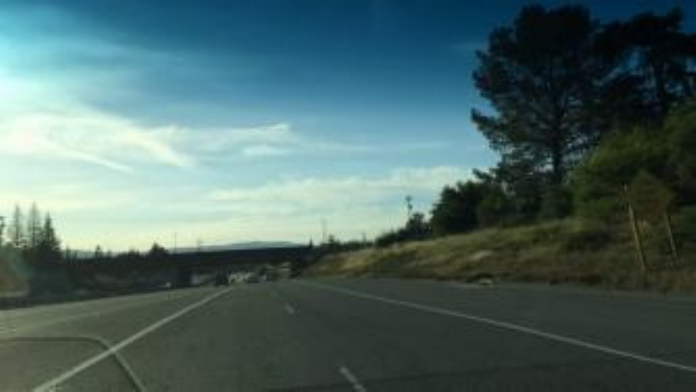} &
\includegraphics[width=0.220\textwidth]{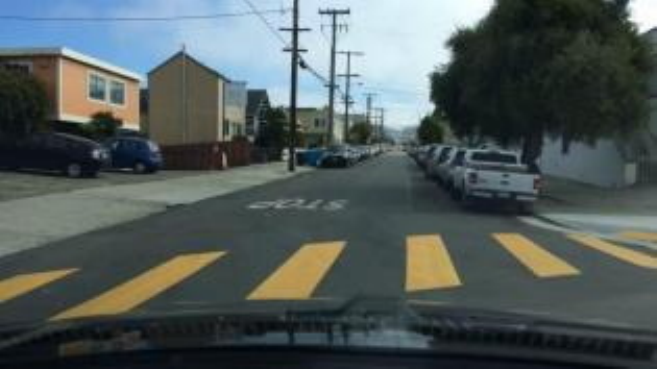} \\
\footnotesize (b) CycleGAN-Turbo &
\includegraphics[width=0.220\textwidth]{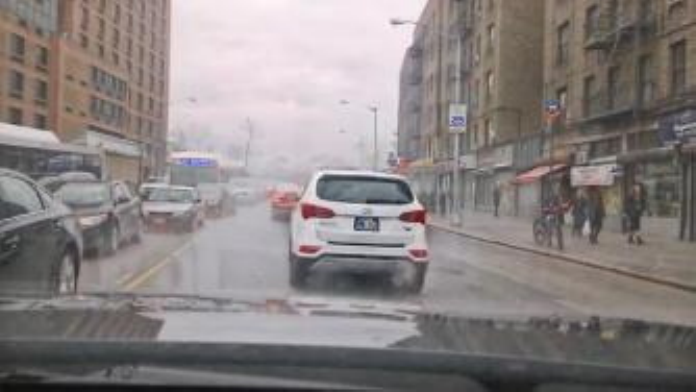} &
\includegraphics[width=0.220\textwidth]{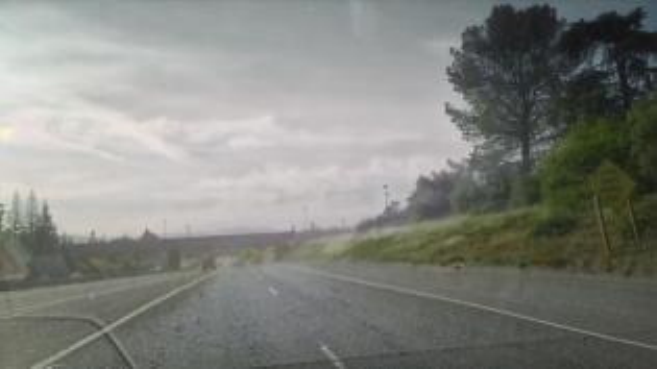} &
\includegraphics[width=0.220\textwidth]{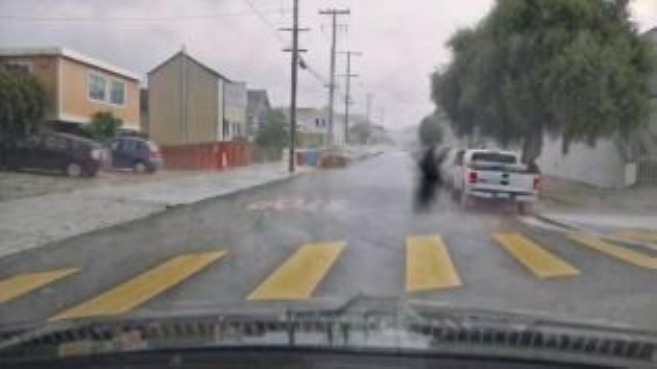} \\
\footnotesize (c) Gemini 3.1 Flash &
\includegraphics[width=0.220\textwidth]{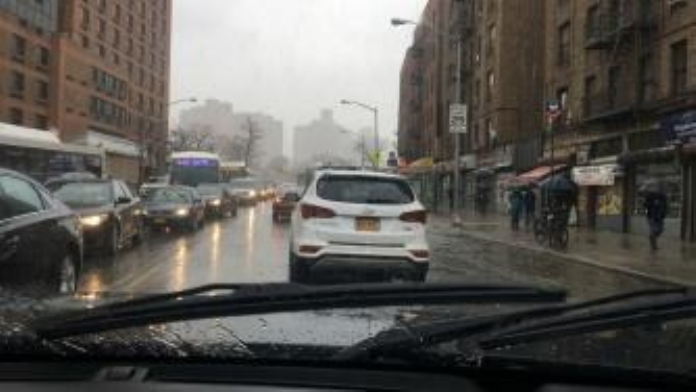} &
\includegraphics[width=0.220\textwidth]{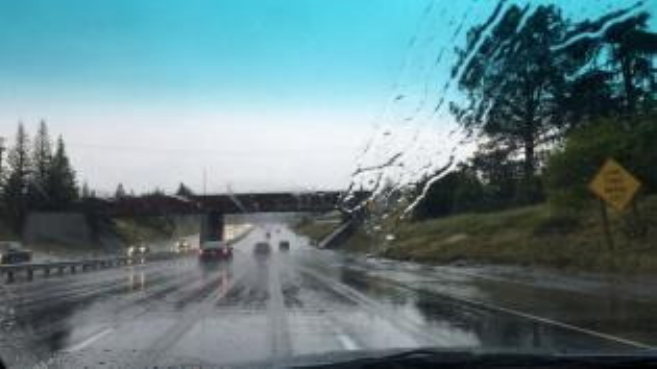} &
\includegraphics[width=0.220\textwidth]{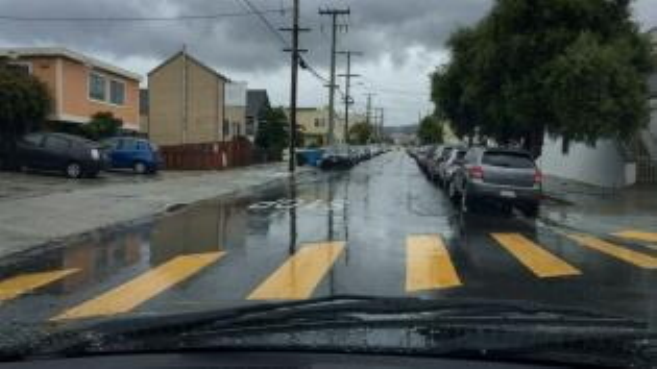} \\
\footnotesize (d) WeatherEdit &
\includegraphics[width=0.220\textwidth]{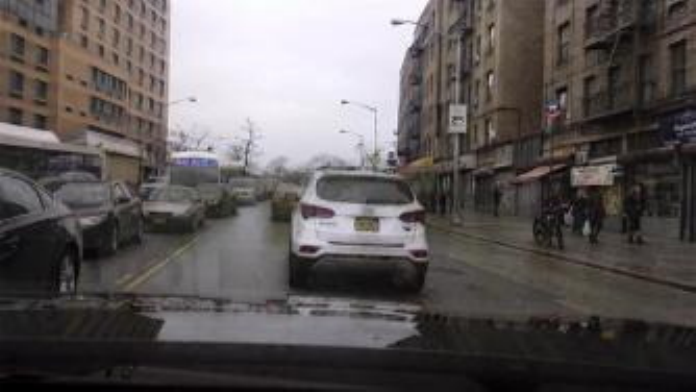} &
\includegraphics[width=0.220\textwidth]{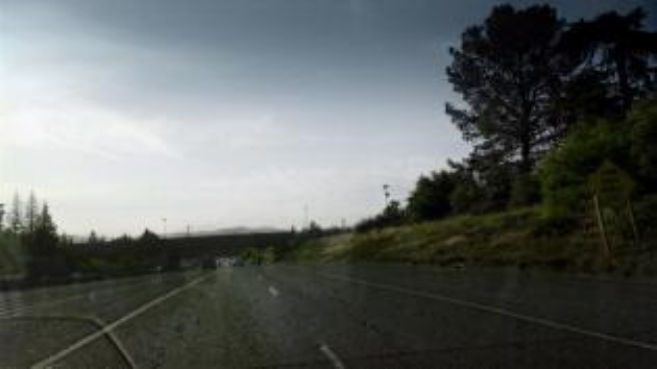} &
\includegraphics[width=0.220\textwidth]{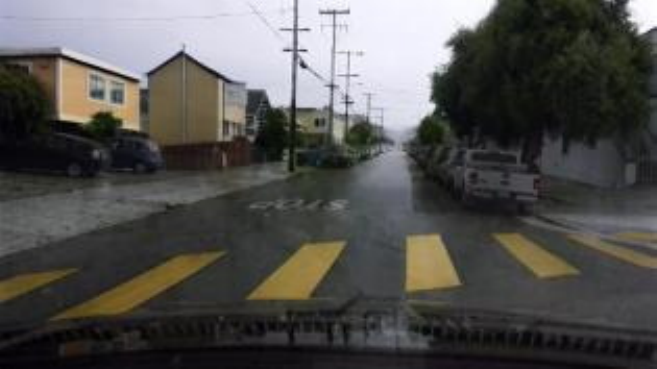} \\
\footnotesize (e) GSRAIN &
\includegraphics[width=0.220\textwidth]{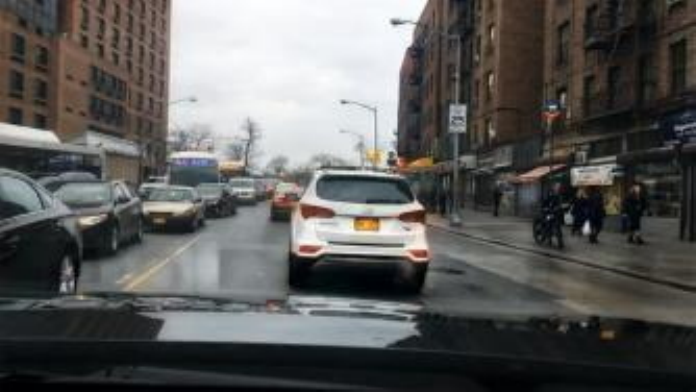} &
\includegraphics[width=0.220\textwidth]{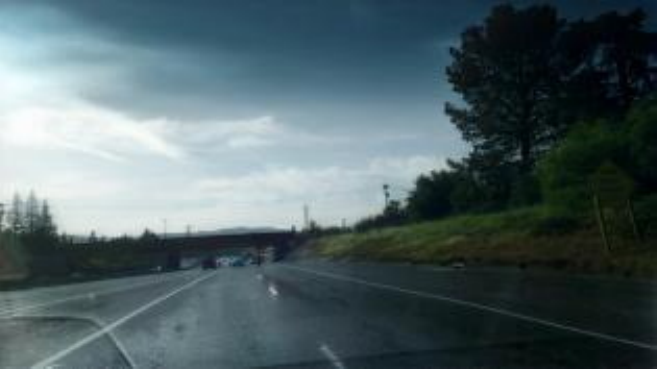} &
\includegraphics[width=0.220\textwidth]{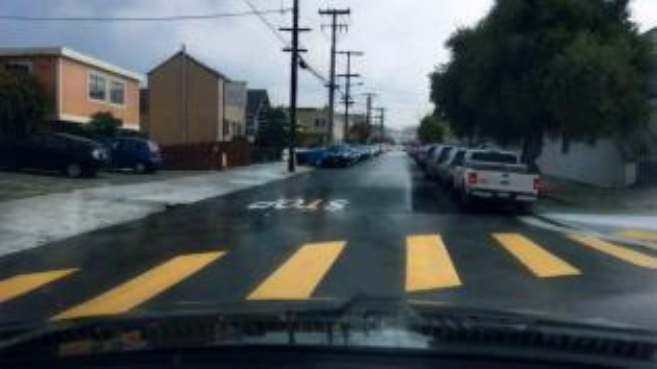}
\end{tabular}
\caption{Style-transfer comparison on BDD-100K. The columns show three driving scenes, and rows (a)--(e) show the clear inputs, CycleGAN-Turbo \cite{ref37}, Gemini 3.1 Flash \cite{ref43}, WeatherEdit \cite{ref25}, and GSRAIN.}
\label{fig:bdd}
\end{figure*}

The BDD-100K single-view results are shown in Fig.~\ref{fig:bdd}. CycleGAN-Turbo produces evident haze-like blur, while rainy features in the sky and buildings remain weak. In the displayed examples, Gemini 3.1 Flash generates a pronounced rainy appearance, but still changes parts of the scene despite a prompt that explicitly requests strict content preservation. WeatherEdit produces a relatively consistent global transfer; however, its changes are mainly reflected in brightness, and the transfer around vehicles and road surfaces remains limited. The displayed GSRAIN results preserve the original scene content more effectively and produce coordinated low-frequency rainy appearance over roads, vehicles, and buildings.

\begin{figure*}[!t]
\centering
\setlength{\tabcolsep}{1.4pt}
\renewcommand{\arraystretch}{0.98}
\begin{tabular}{@{}>{\raggedleft\arraybackslash}m{0.12\textwidth}@{\hspace{3pt}}ccc@{}}
\footnotesize (a) Clear input &
\includegraphics[width=0.210\textwidth]{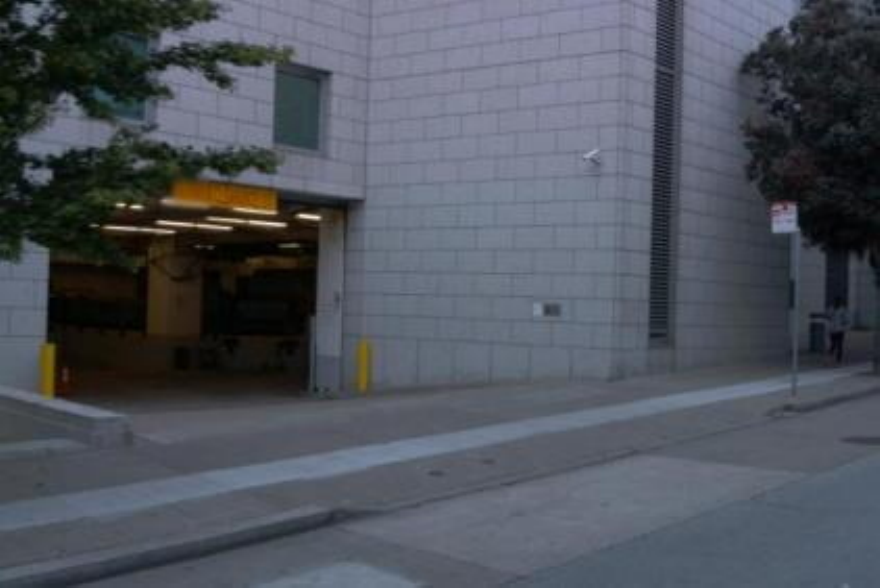} &
\includegraphics[width=0.210\textwidth]{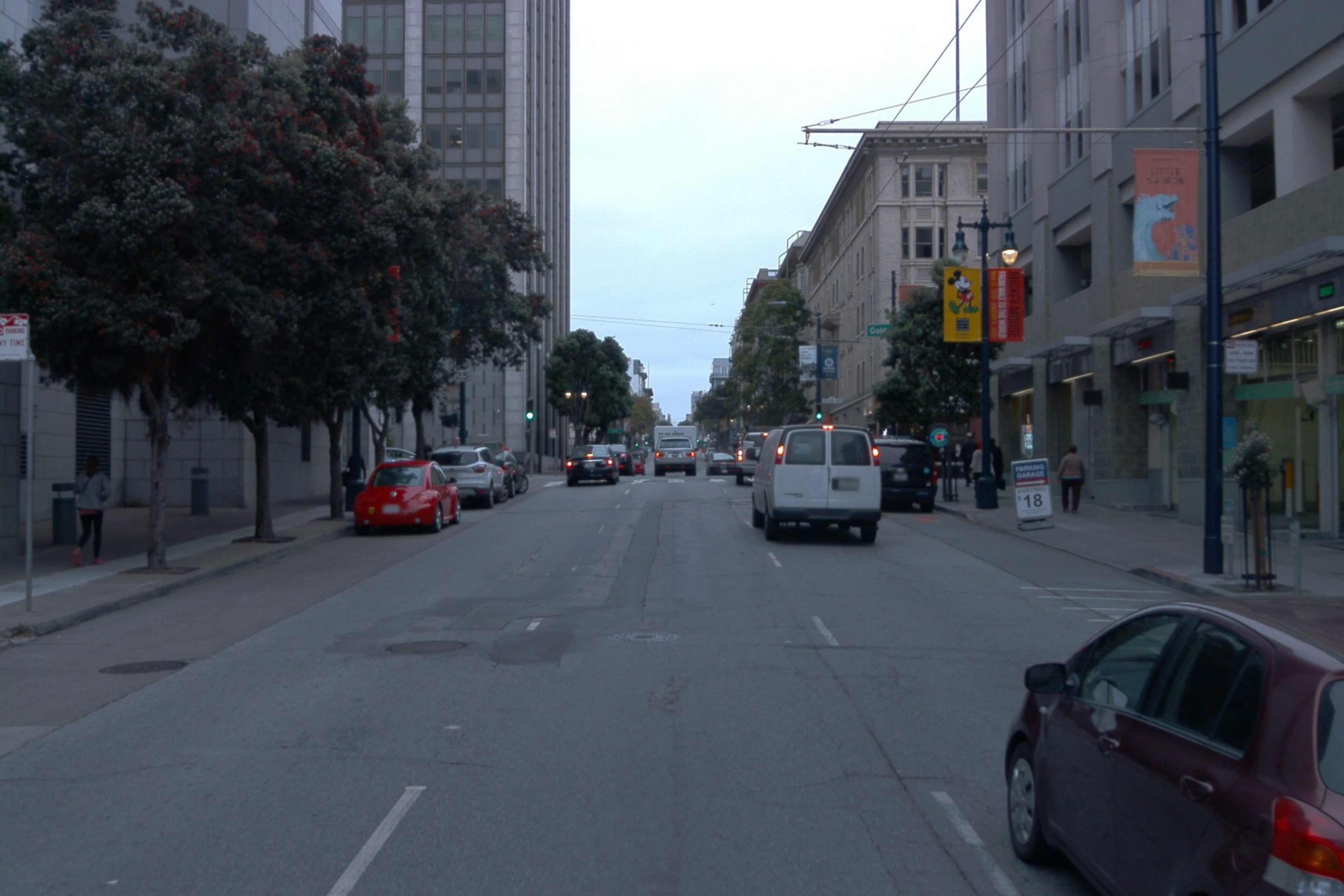} &
\includegraphics[width=0.210\textwidth]{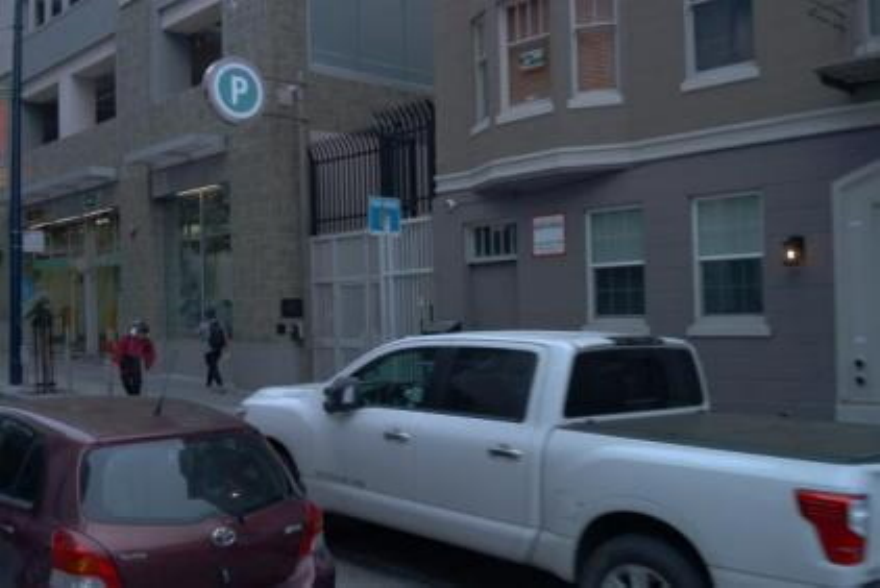} \\
\footnotesize (b) CycleGAN-Turbo &
\includegraphics[width=0.210\textwidth]{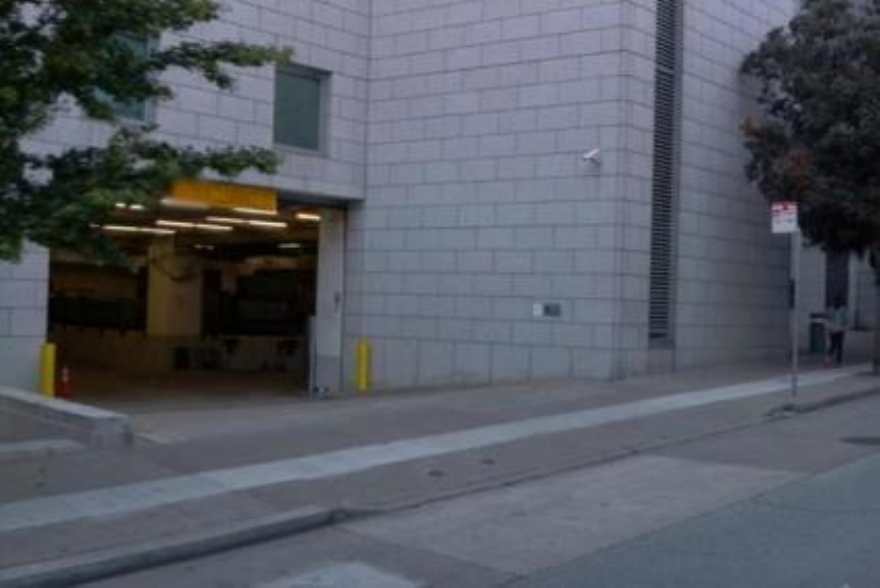} &
\includegraphics[width=0.210\textwidth]{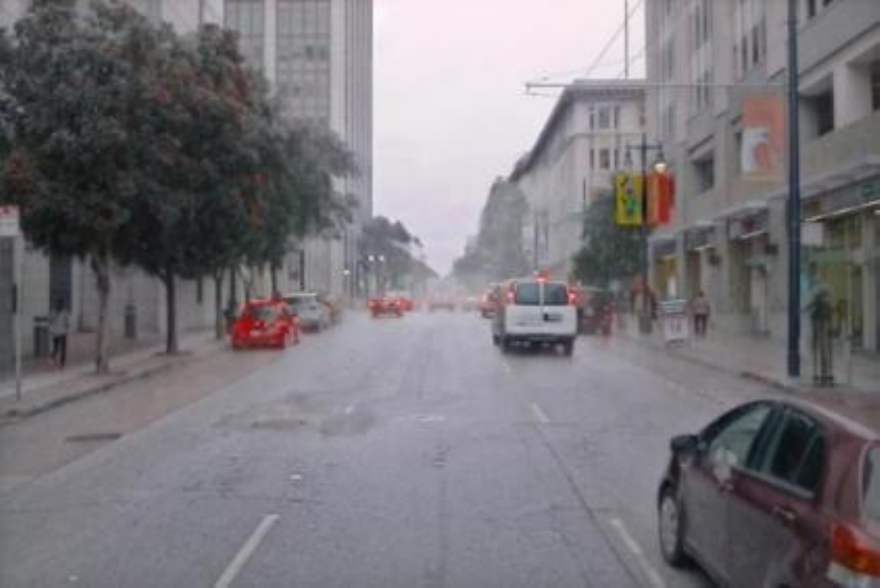} &
\includegraphics[width=0.210\textwidth]{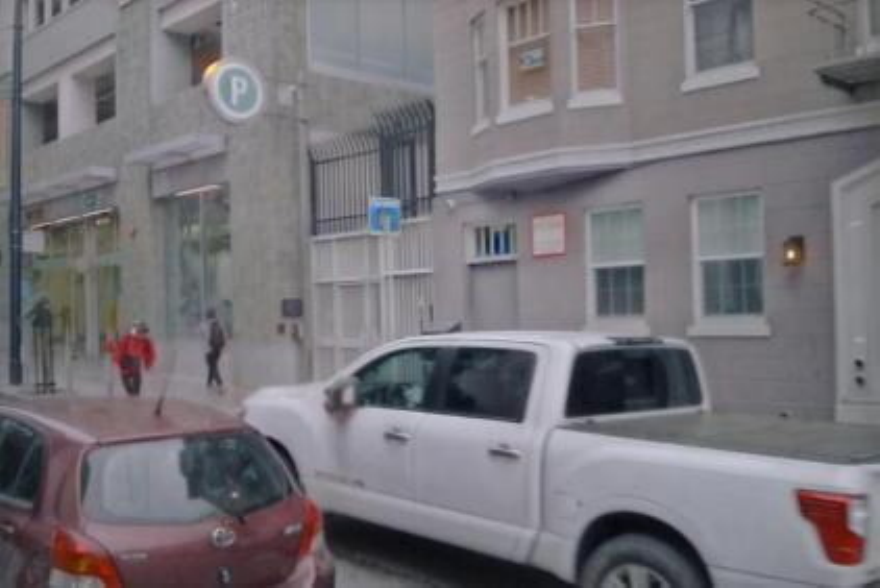} \\
\footnotesize (c) Gemini 3.1 Flash &
\includegraphics[width=0.210\textwidth]{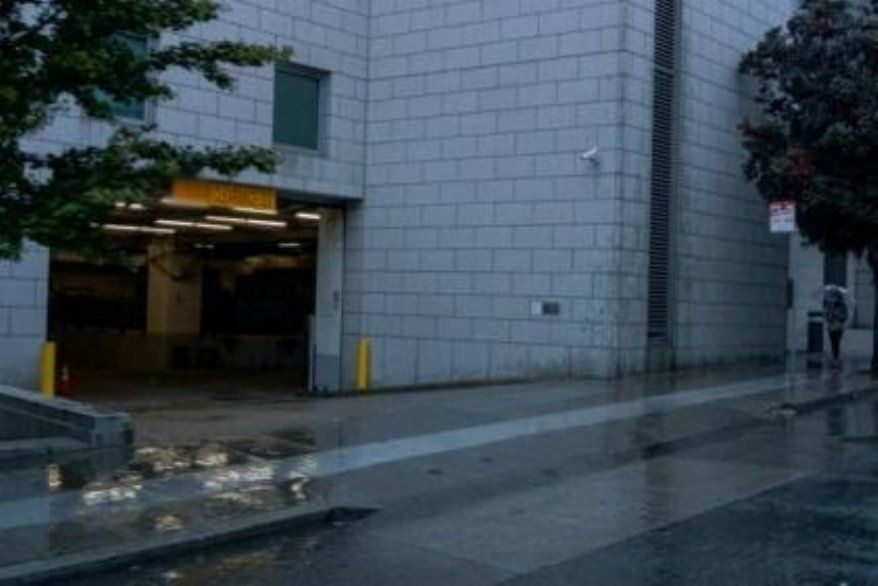} &
\includegraphics[width=0.210\textwidth]{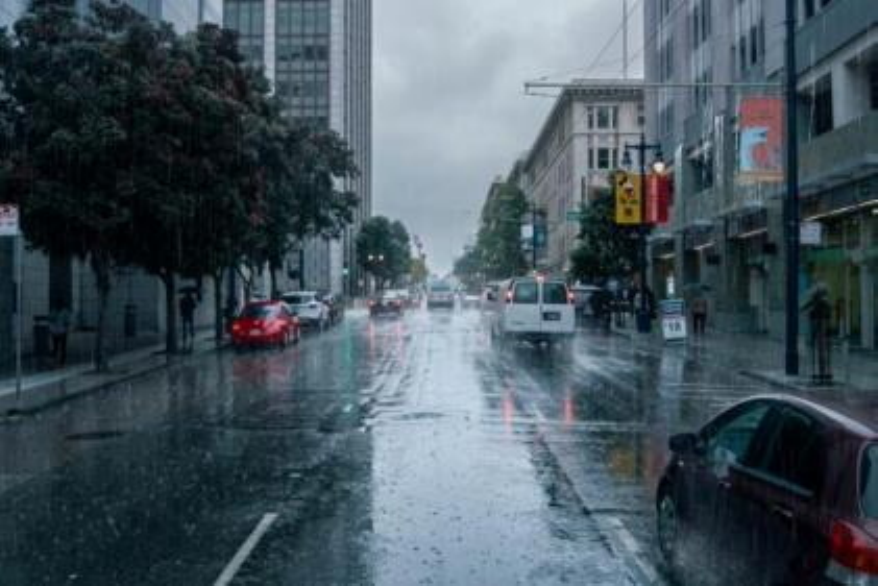} &
\includegraphics[width=0.210\textwidth]{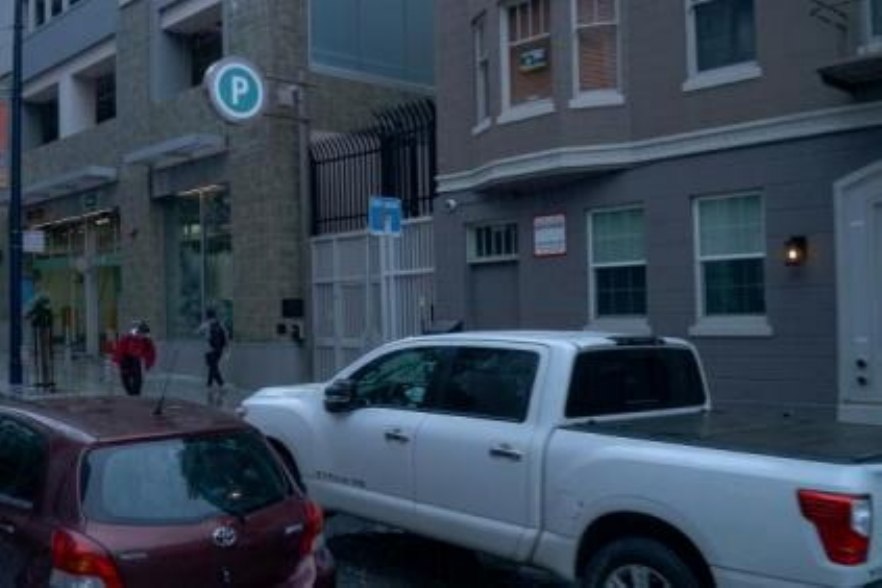} \\
\footnotesize (d) WeatherEdit &
\includegraphics[width=0.210\textwidth]{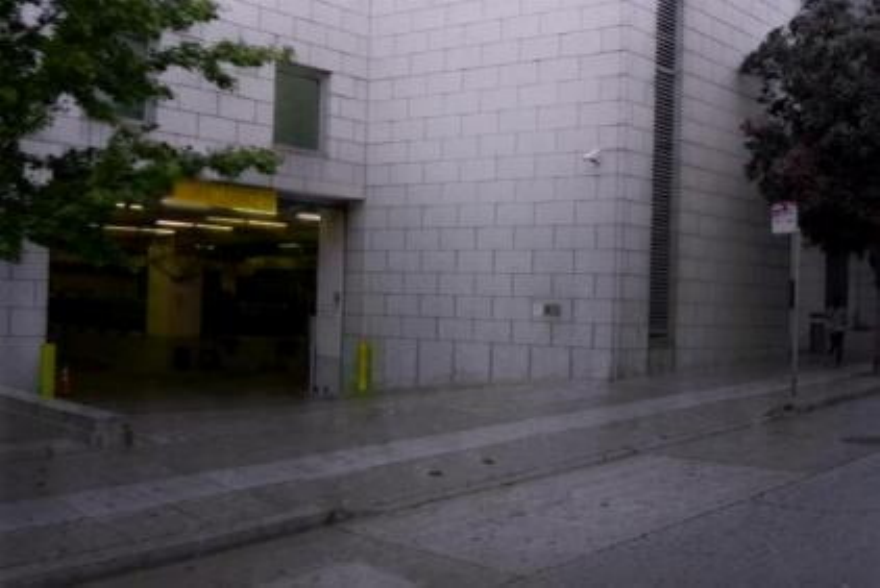} &
\includegraphics[width=0.210\textwidth]{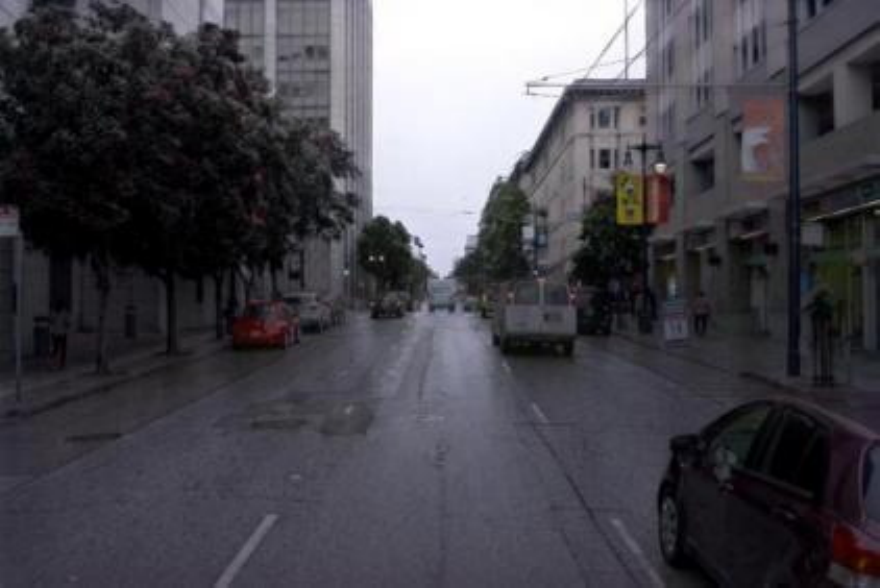} &
\includegraphics[width=0.210\textwidth]{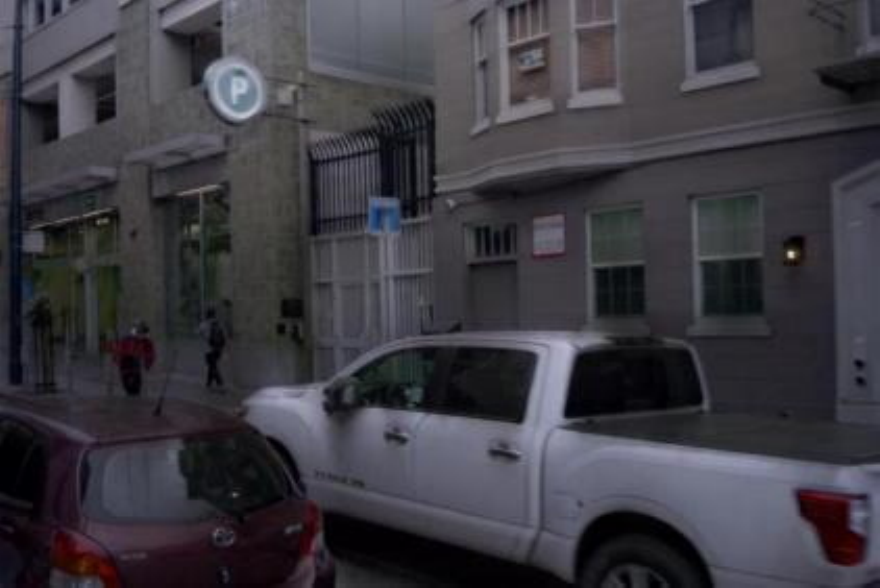} \\
\footnotesize (e) GSRAIN &
\includegraphics[width=0.210\textwidth]{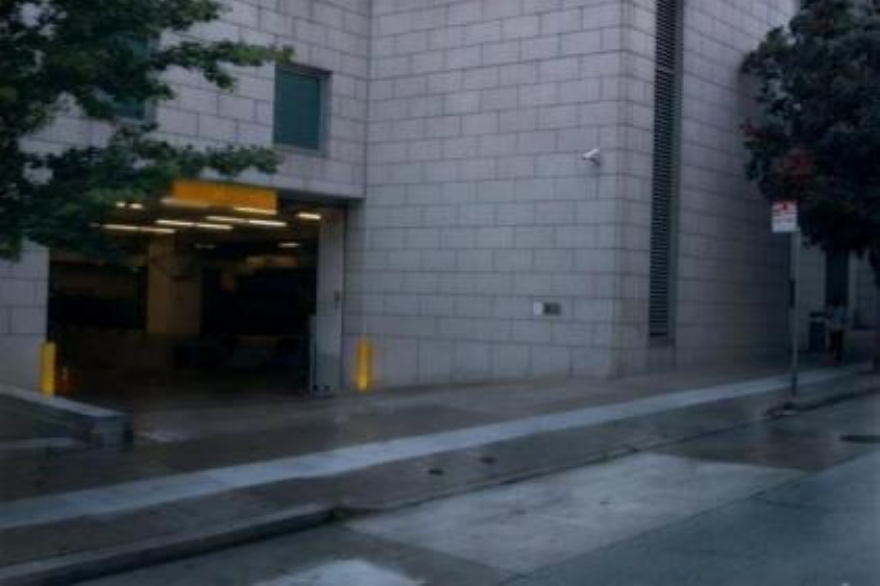} &
\includegraphics[width=0.210\textwidth]{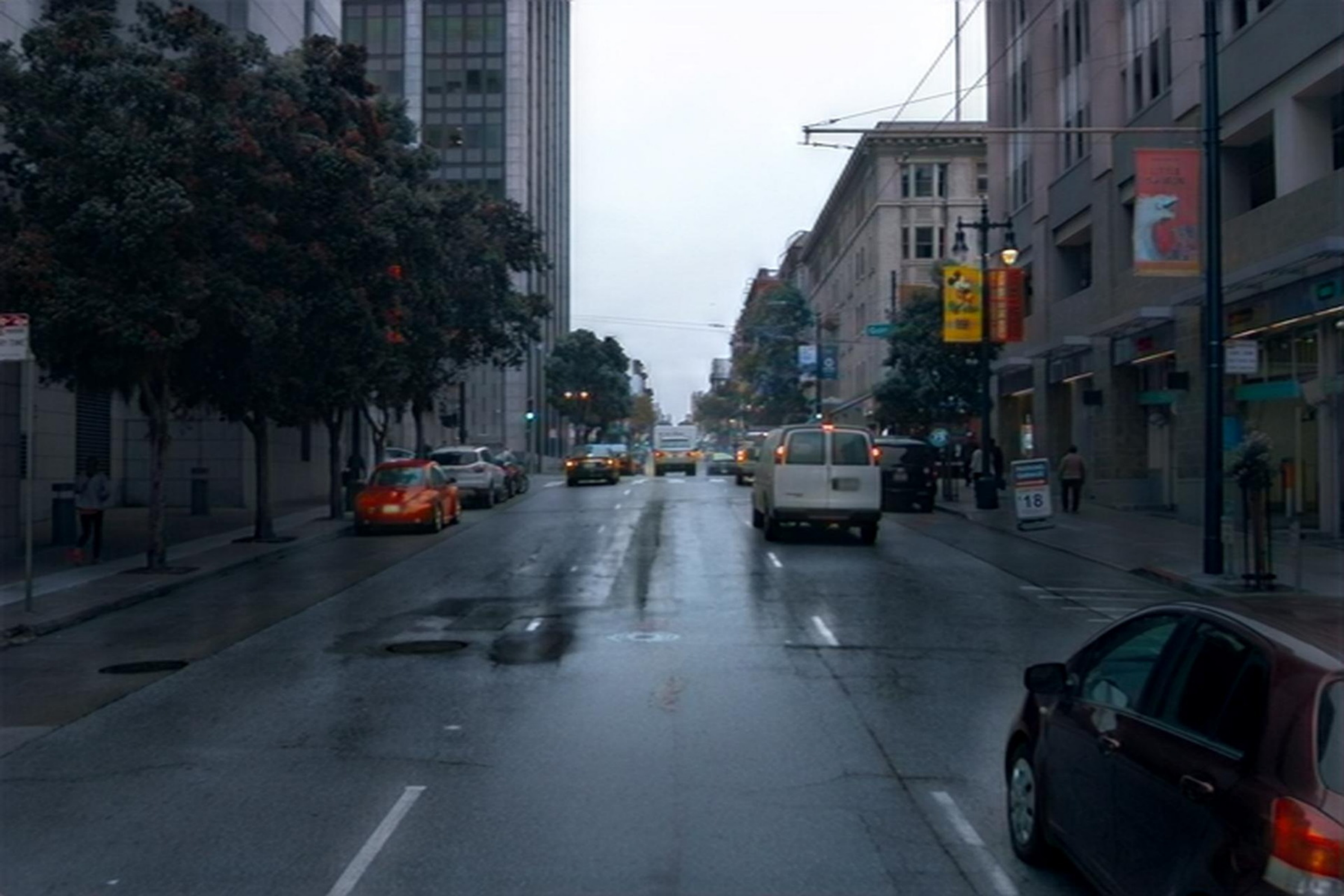} &
\includegraphics[width=0.210\textwidth]{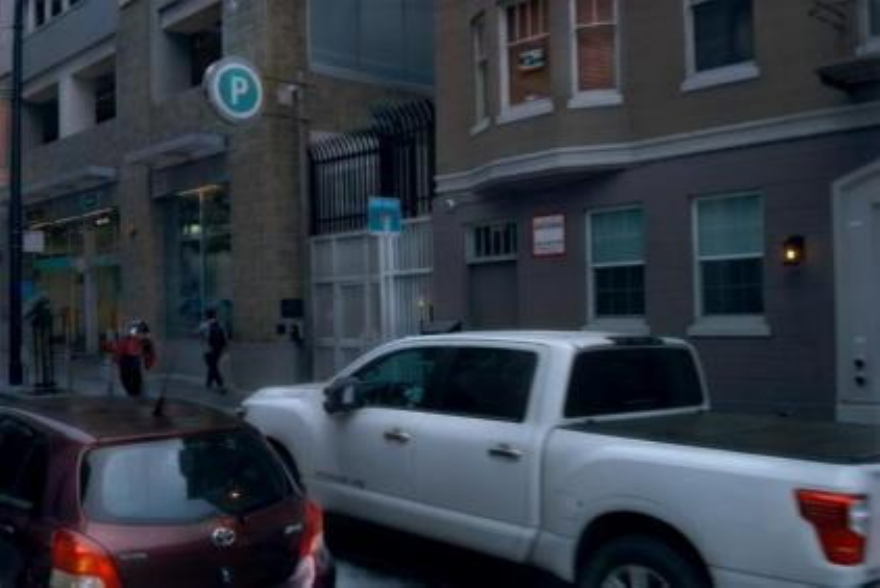}
\end{tabular}
\caption{Multi-view style-transfer comparison on Waymo. The columns correspond to the selected left, center, and right camera views.}
\label{fig:waymo}
\end{figure*}

Figure~\ref{fig:waymo} presents multi-view transfer results on the Waymo dataset \cite{ref44}, using three forward camera views (indices 0--2). CycleGAN-Turbo mainly changes global brightness and lacks characteristic rainy cues such as wet ground, reflections, and an overall dim appearance. Its left-view transfer is also weaker than those of the other views. In the displayed sample, Gemini 3.1 Flash generates a pronounced rainy appearance but changes scene content differently across views; for example, an umbrella is added above a pedestrian only in the left view. WeatherEdit maintains relatively coordinated global transfer, but its changes are dominated by brightness, and the rainy appearance of vehicles, buildings, and lights remains limited. In the shown multi-view examples, GSRAIN better preserves the main content and scene structure while providing a more coordinated cross-view rainy appearance, thereby supplying more consistent inputs for subsequent 3DGS reconstruction.

\subsubsection{Quantitative Comparison}
\label{subsubsec:quantitative}

Fr\'{e}chet Inception Distance (FID) \cite{ref45} measures the discrepancy between the feature distributions of generated and real images; a lower value indicates closer similarity to real rainy images. Under the same dataset and training protocol, GSRAIN obtains the lowest FID of 149.09, compared with 155.71 for CycleGAN-Turbo and 157.94 for WeatherEdit, as shown in Table~\ref{tab:fid}. Gemini 3.1 Flash is accessed through a pretrained model API and cannot be reproduced under the same training protocol, so it is included only in the qualitative comparison.

\begin{table}[!t]
\caption{FID Comparison}
\label{tab:fid}
\centering
\footnotesize
\begin{tabular}{lc}
\toprule
Method & FID $\downarrow$ \\
\midrule
CycleGAN-Turbo \cite{ref37} & 155.71 \\
WeatherEdit \cite{ref25} & 157.94 \\
GSRAIN & \textbf{149.09} \\
\bottomrule
\end{tabular}
\end{table}

\subsubsection{Ablation Study}
\label{subsubsec:ablation}

\begin{figure*}[!t]
\centering
\subfloat[Clear input]{\includegraphics[width=0.275\textwidth]{figures/image21.pdf}}\hfil
\subfloat[RGB only]{\includegraphics[width=0.275\textwidth]{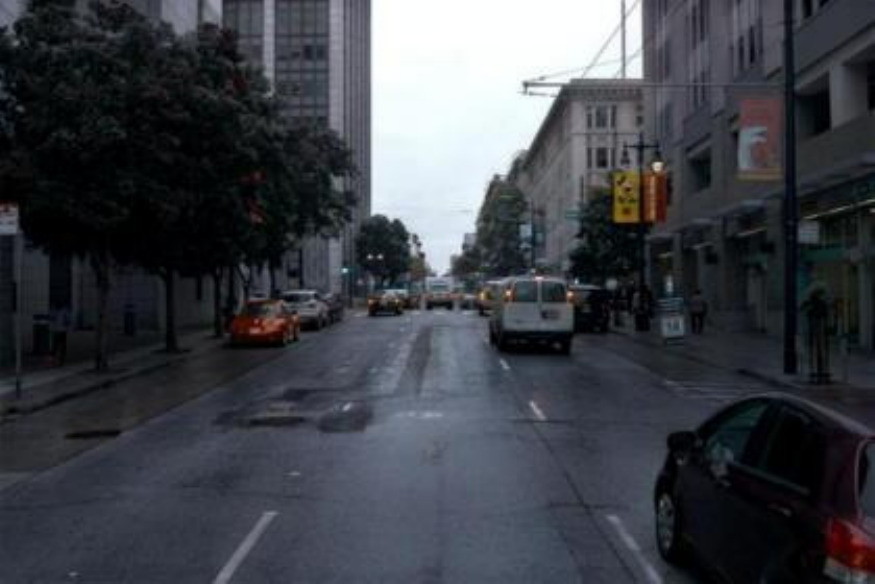}}\hfil
\subfloat[RGB and normal map]{\includegraphics[width=0.275\textwidth]{figures/image33.pdf}}
\caption{Ablation study of the normal-map channel.}
\label{fig:normal-ablation}
\end{figure*}

Figure~\ref{fig:normal-ablation} compares RGB-only input with RGB plus the normal-map channel. In the displayed example, adding the normal condition produces more evident wet reflections in ground regions where the surface orientation is clear. This observation suggests that the normal-map condition helps the model use ground-surface orientation and improves the appearance representation of wet reflective regions.

\begin{figure*}[!t]
\centering
\subfloat[Clear input]{\includegraphics[width=0.275\textwidth]{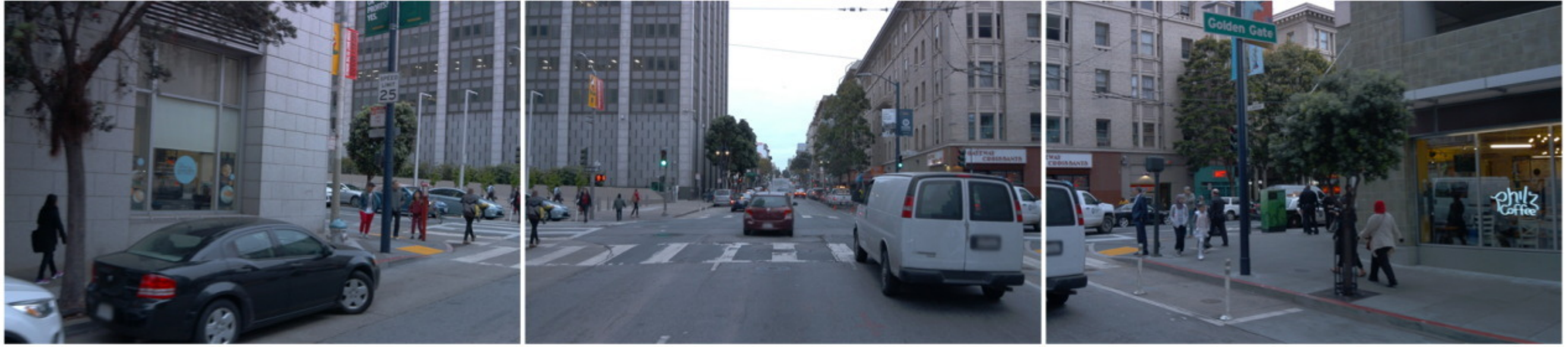}}\hfil
\subfloat[Without mutual self-attention]{\includegraphics[width=0.275\textwidth]{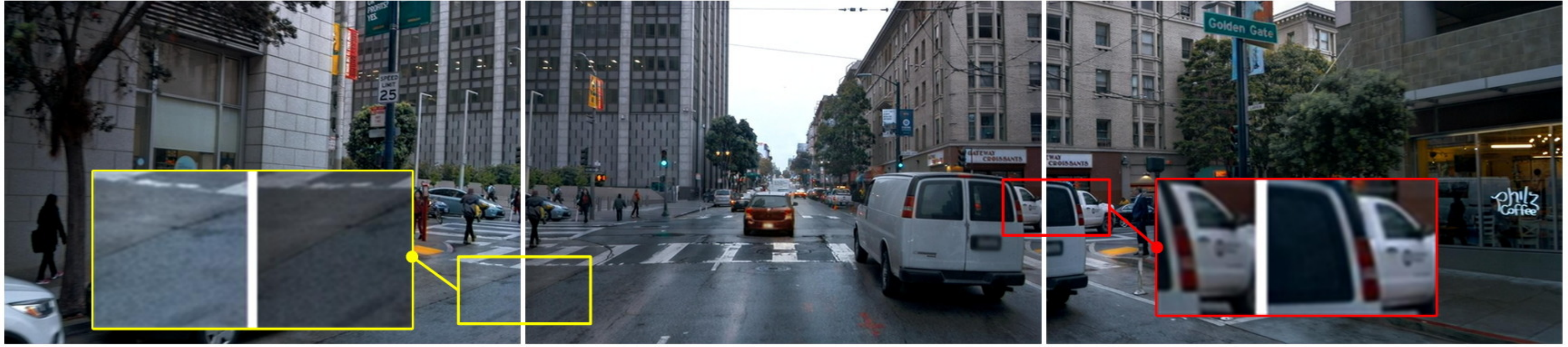}}\hfil
\subfloat[With mutual self-attention]{\includegraphics[width=0.275\textwidth]{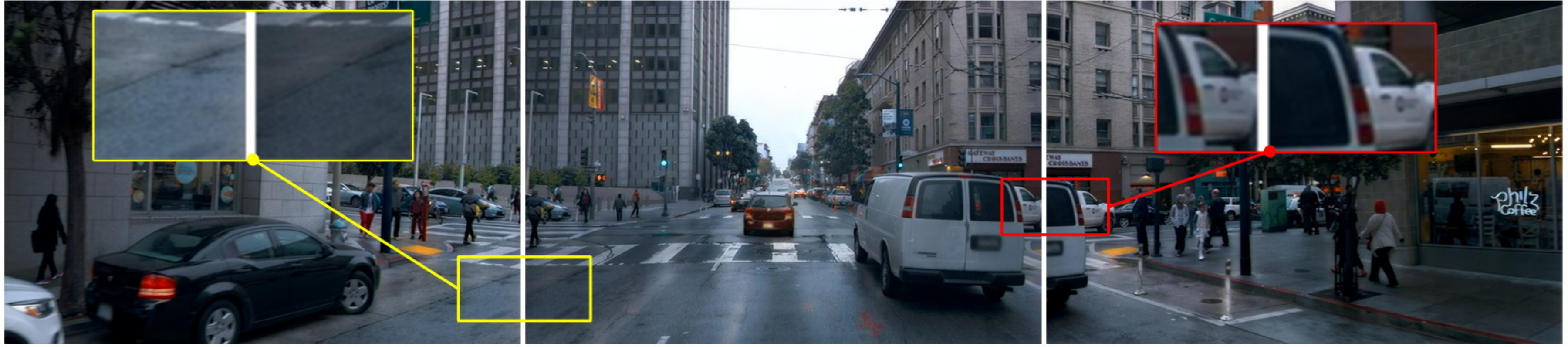}}
\caption{Ablation study of mutual self-attention.}
\label{fig:attention-ablation}
\end{figure*}

Figure~\ref{fig:attention-ablation} compares mutual self-attention with standard self-attention in the selected multi-view example. With mutual self-attention, the road-color difference between the left and center views is reduced, and the vehicle brightness in the right view becomes closer to that in the center view. The displayed result suggests that mutual self-attention improves the coordination of rainy appearance across views.

\subsection{Overall Rainfall Synthesis}
\label{subsec:overall-rain}

\subsubsection{Fusion of High- and Low-Frequency Effects}
\label{subsubsec:fusion}

The low-frequency branch transfers geometry-constrained rainy appearance and reconstructs or re-optimizes a rainy 3DGS background. The high-frequency branch fits raindrop diameter, velocity, and number density from measured rainfall data and produces near-field rain-streak and far-field haze Gaussians. During rendering, all three types of Gaussians participate in depth-ordered alpha compositing, unifying global rainy appearance and local raindrop disturbances in one 3D Gaussian representation.

\subsubsection{Experimental Setup}
\label{subsubsec:overall-setup}

After diffusion-model fine-tuning, 3DGS scene re-optimization and high-frequency rainfall simulation are conducted on the Waymo dataset using one NVIDIA RTX 4090 GPU. The image resolution is $960\times640$ pixels. The rainfall intensity is set to 8.5~mm/h. The gamma shape and scale parameters are 7.132 and 0.114~mm, respectively, and the raindrop number density is 1190~m$^{-3}$. The critical haze coefficient is $\varphi=2$. Since raindrop diameters under this intensity are generally below 1.5~mm, \eqref{eq:visible-distance} gives a maximum visible distance of approximately 3~m. In the ego-vehicle coordinate system, the rain-streak region is $[-1,5]\times[-3,3]\times[-1.5,4.5]$~m, and the haze region is $[-10,60]\times[-10,10]\times[-1.5,20.5]$~m. Haze opacity is 0.05 and the exposure time is 0.05~s.

\subsubsection{Complete-Effect Comparison}
\label{subsubsec:complete-comparison}

\begin{figure}[!t]
\centering
\setlength{\tabcolsep}{0pt}
\begin{tabular}{c}
\includegraphics[width=0.96\linewidth]{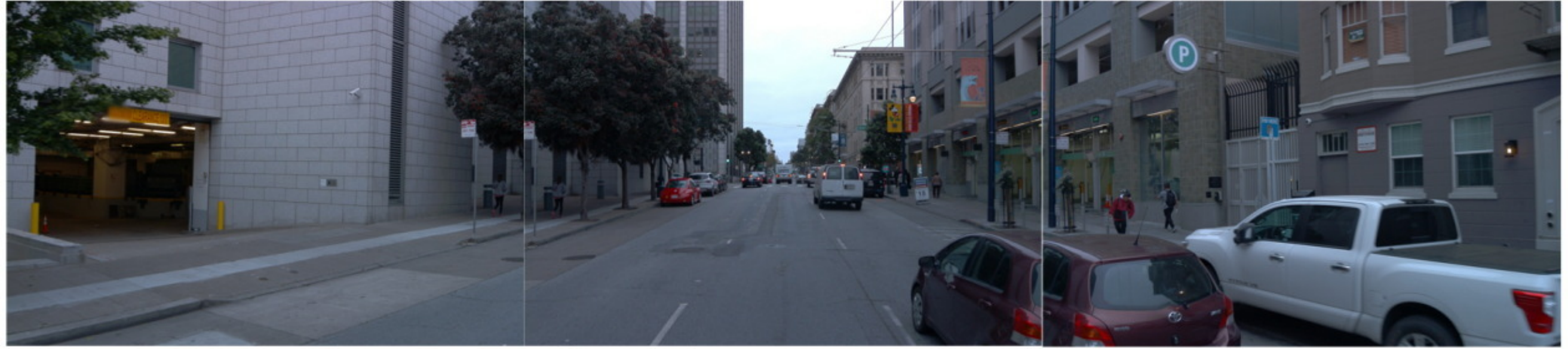}\\[-1pt]
{\footnotesize (a) Clear input}\\[1.5pt]
\includegraphics[width=0.96\linewidth]{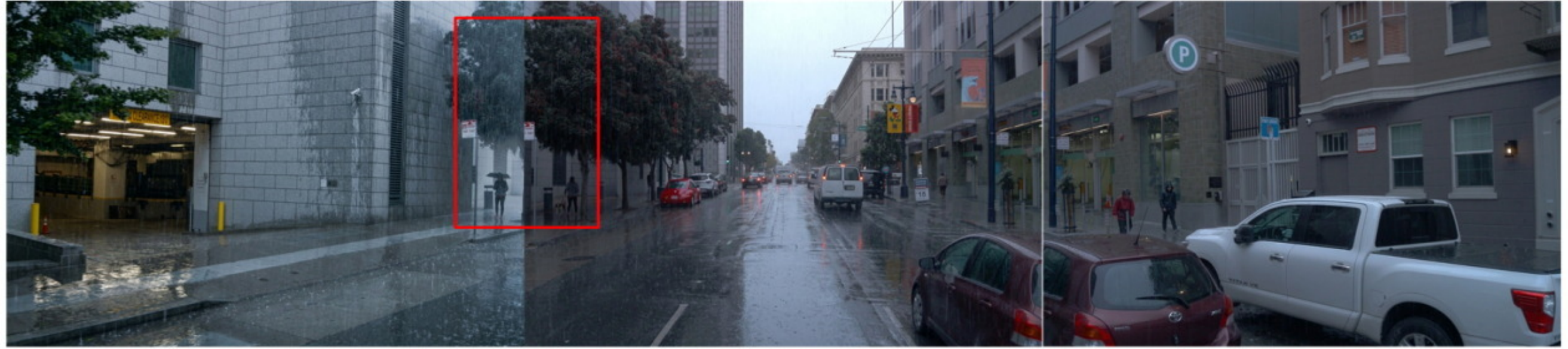}\\[-1pt]
{\footnotesize (b) Gemini 3.1 Flash~\cite{ref43}}\\[1.5pt]
\includegraphics[width=0.96\linewidth]{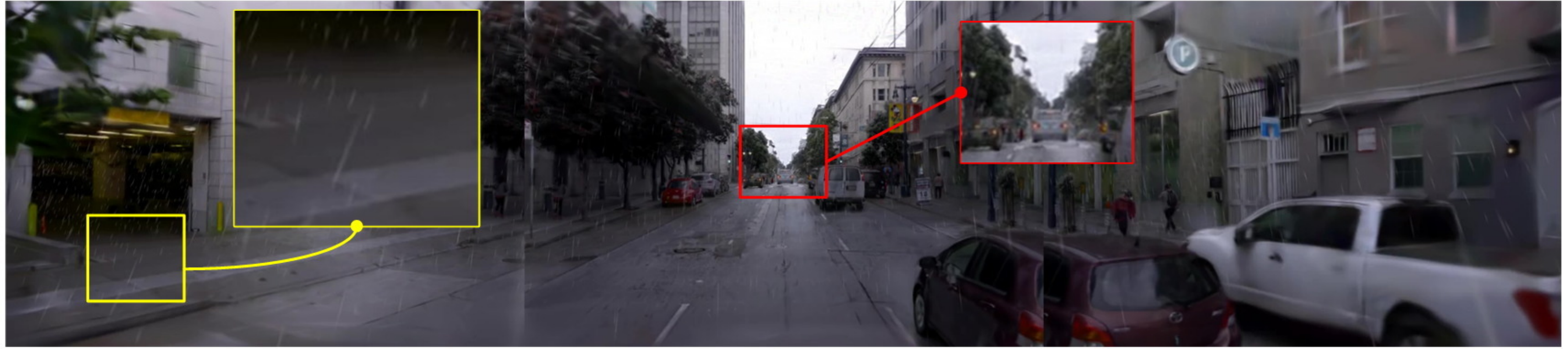}\\[-1pt]
{\footnotesize (c) WeatherEdit~\cite{ref25}}\\[1.5pt]
\includegraphics[width=0.96\linewidth]{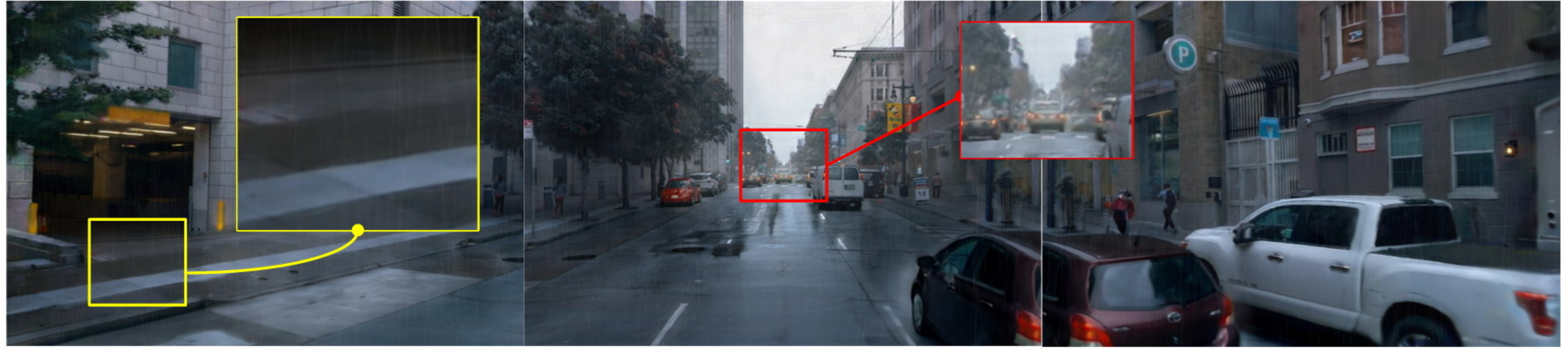}\\[-1pt]
{\footnotesize (d) GSRAIN}
\end{tabular}
\caption{Qualitative comparison of complete rainfall synthesis. Gemini 3.1 Flash changes content and lacks numerical intensity control; WeatherEdit does not show a far-field haze component; GSRAIN jointly represents low-frequency rainy appearance and calibrated rain-streak/haze effects. Red and yellow boxes highlight vehicle and road regions for visual inspection.}
\label{fig:overall-comparison}
\end{figure}

Figure~\ref{fig:overall-comparison} compares complete rainfall effects. Gemini 3.1 Flash generates pronounced rainy appearance from prompts, but changes multi-view content and does not support numerical rainfall-intensity control. WeatherEdit preserves more scene content in the selected sample, but its high-frequency component is mainly limited to near-field rain streaks and does not show the far-field haze modeled by GSRAIN. GSRAIN jointly renders the low-frequency rainy background, near-field rain streaks, and far-field haze, with explicit mappings from high-frequency parameters to measured rainfall statistics.

The selected views show that GSRAIN preserves the main multi-view content while simultaneously producing global rainy appearance, near-field streaks, and far-field haze. This demonstrates the coordinated representation of high- and low-frequency effects in one 3DGS scene.

\subsubsection{Rainfall-Intensity Control}
\label{subsubsec:intensity}

GSRAIN adjusts rainfall over 0--13~mm/h through calibrated parameter mappings. Table~\ref{tab:intensity} lists the number density and gamma-distribution parameters for three target intensities. The simulated results in Fig.~\ref{fig:intensity-comparison} show that both rain-streak density and far-field haze increase with rainfall intensity, indicating graded control of the visual rainfall strength.

\begin{table}[!t]
\caption{Raindrop Parameters under Different Rainfall Intensities}
\label{tab:intensity}
\centering
\footnotesize
\setlength{\tabcolsep}{3pt}
\begin{tabular}{cccc}
\toprule
$R$ (mm/h) & Density (m$^{-3}$) & Shape & Scale (mm) \\
\midrule
1.2 & 271 & 9.705 & 0.074 \\
8.5 & 1190 & 7.132 & 0.114 \\
12.0 & 1710 & 10.466 & 0.085 \\
\bottomrule
\end{tabular}
\end{table}

\begin{figure*}[!tb]
\centering
\setlength{\tabcolsep}{2pt}
\begin{tabular}{cc}
\includegraphics[width=0.40\textwidth]{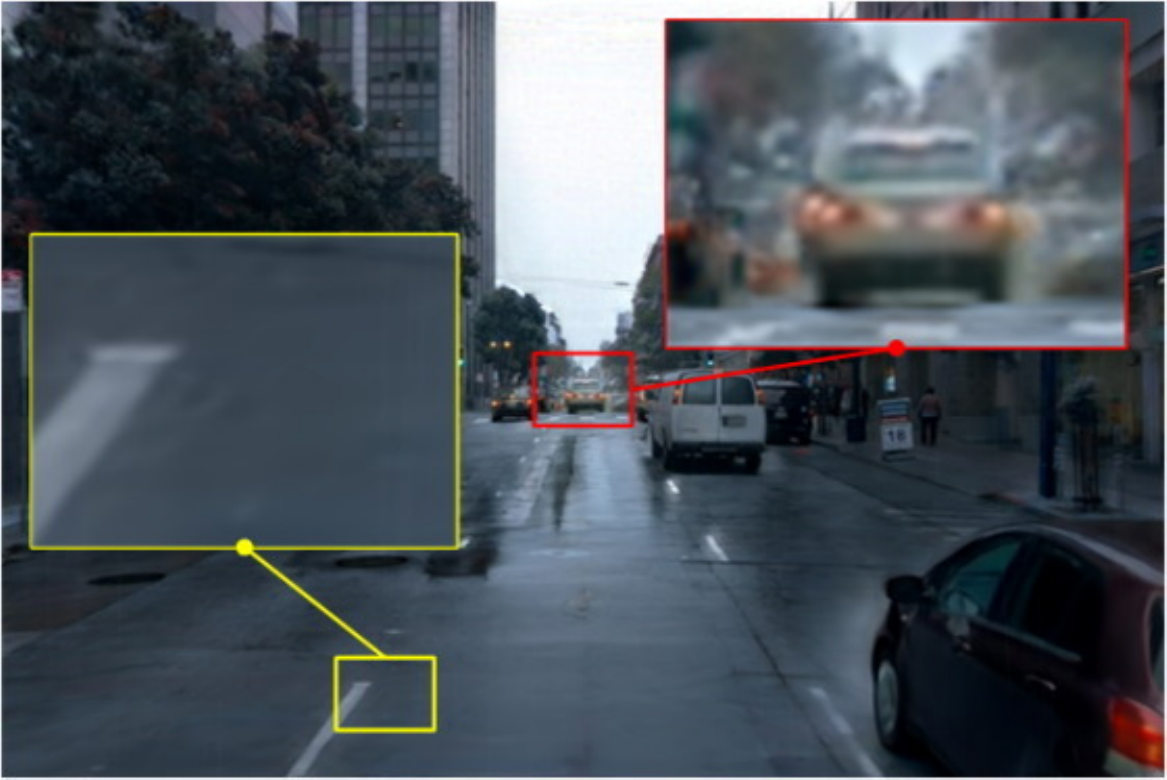} &
\includegraphics[width=0.40\textwidth]{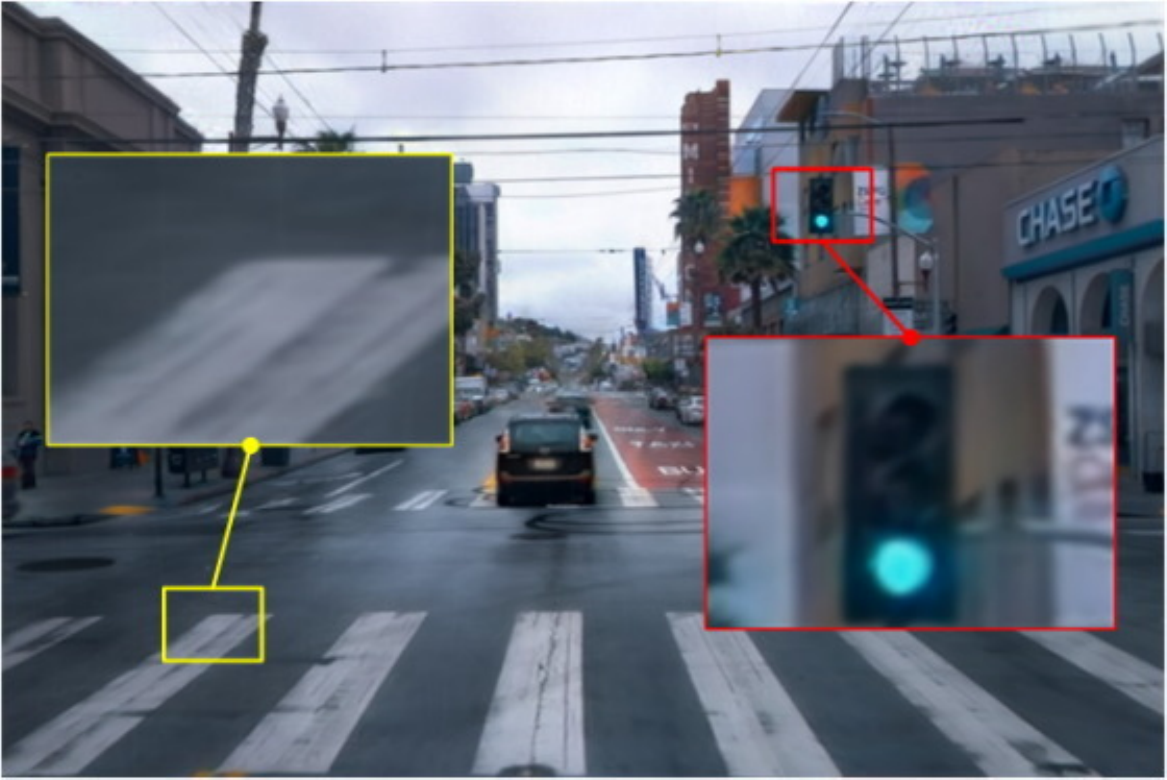}\\[-1pt]
\multicolumn{2}{c}{\footnotesize (a) $R=1.2~\mathrm{mm/h}$}\\[3pt]
\includegraphics[width=0.40\textwidth]{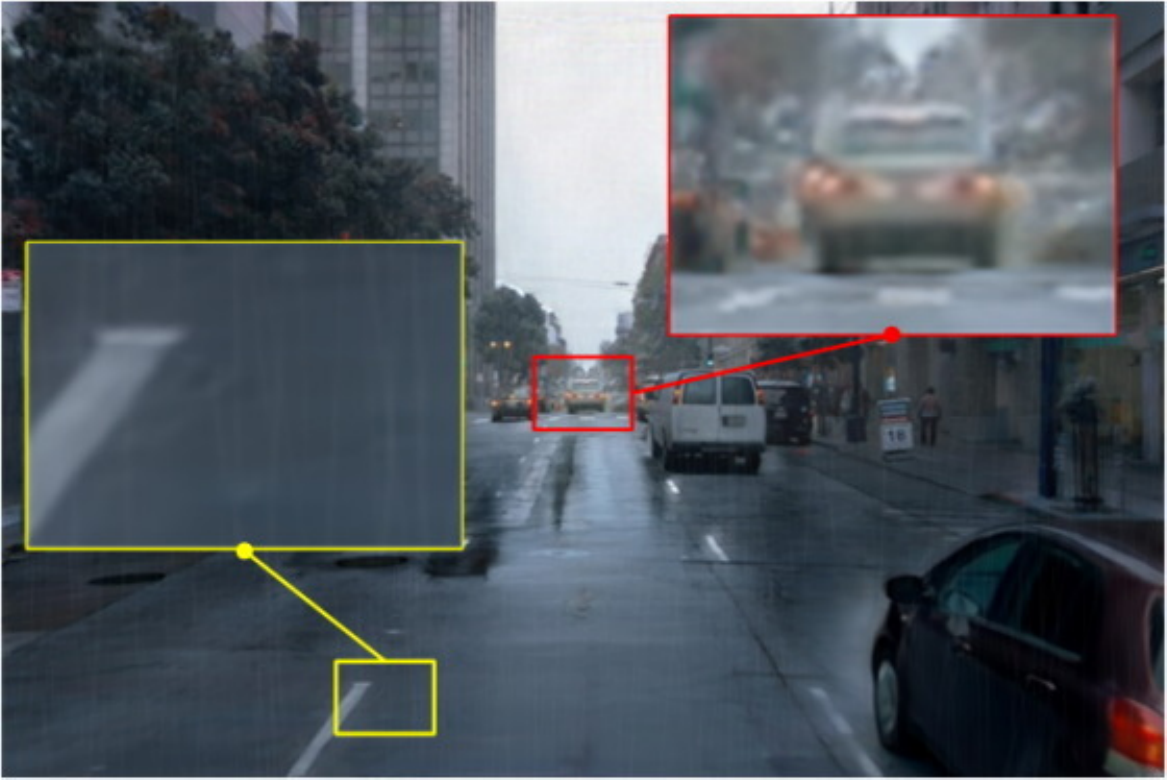} &
\includegraphics[width=0.40\textwidth]{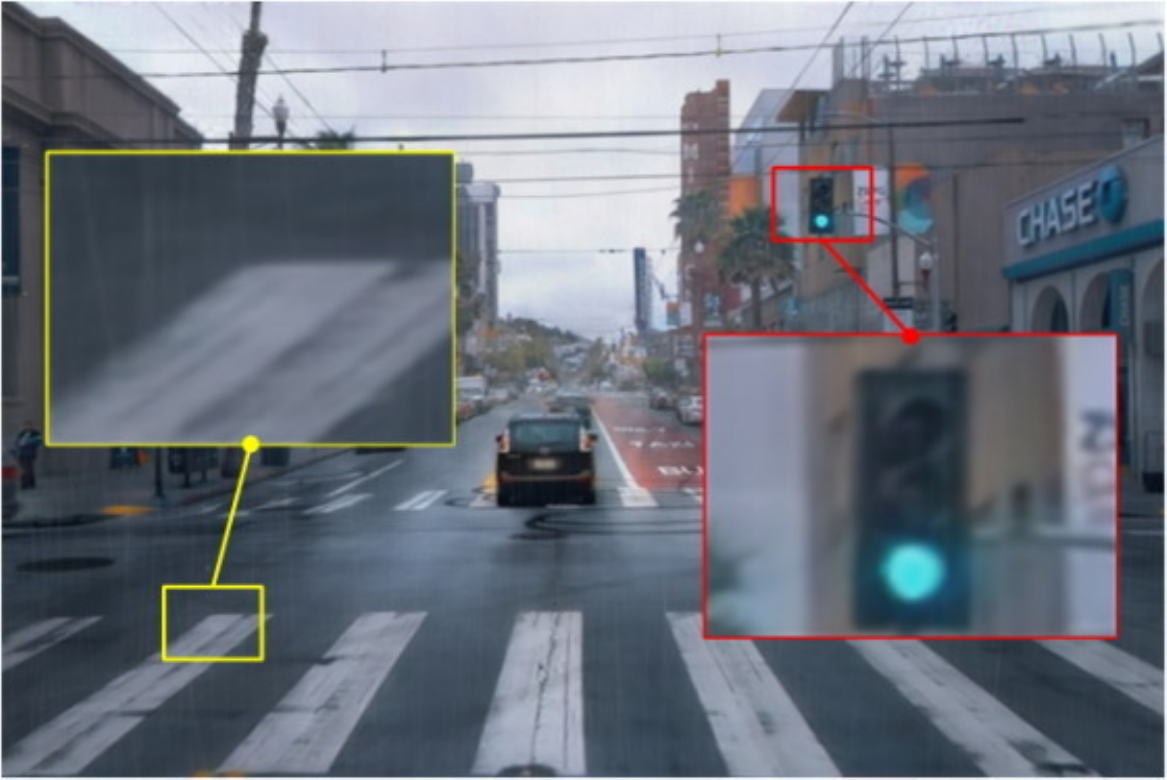}\\[-1pt]
\multicolumn{2}{c}{\footnotesize (b) $R=8.5~\mathrm{mm/h}$}\\[3pt]
\includegraphics[width=0.40\textwidth]{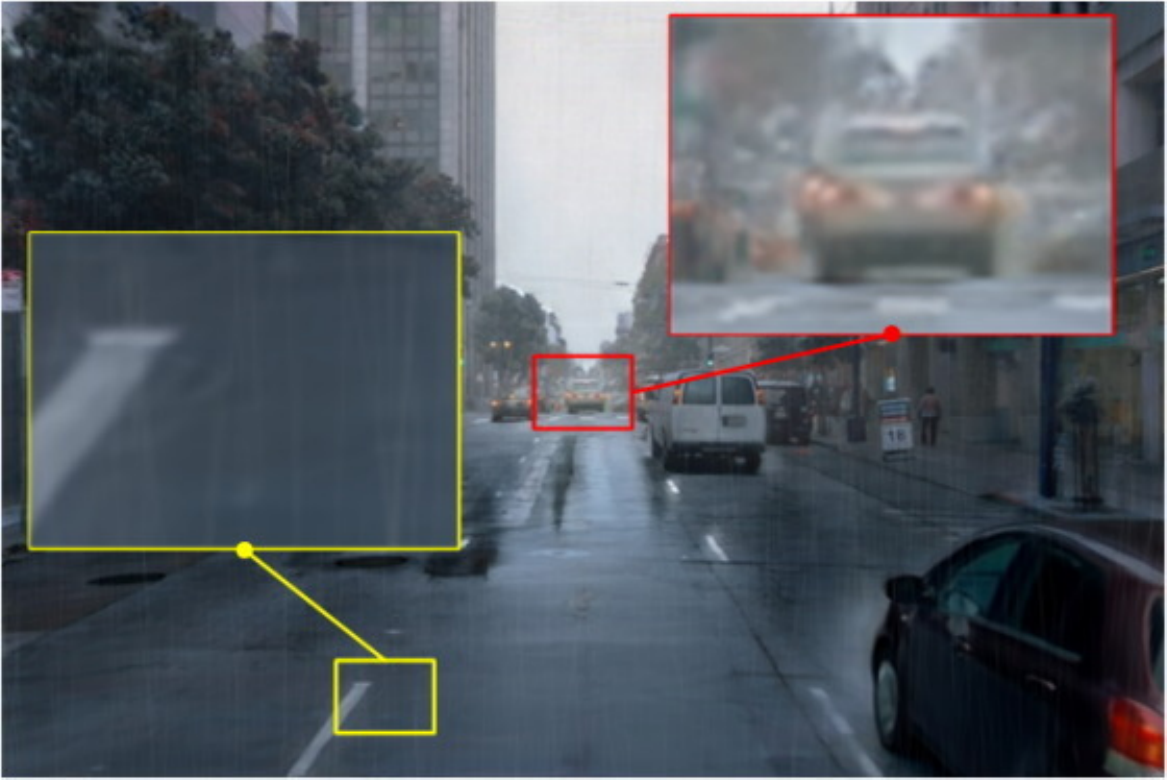} &
\includegraphics[width=0.40\textwidth]{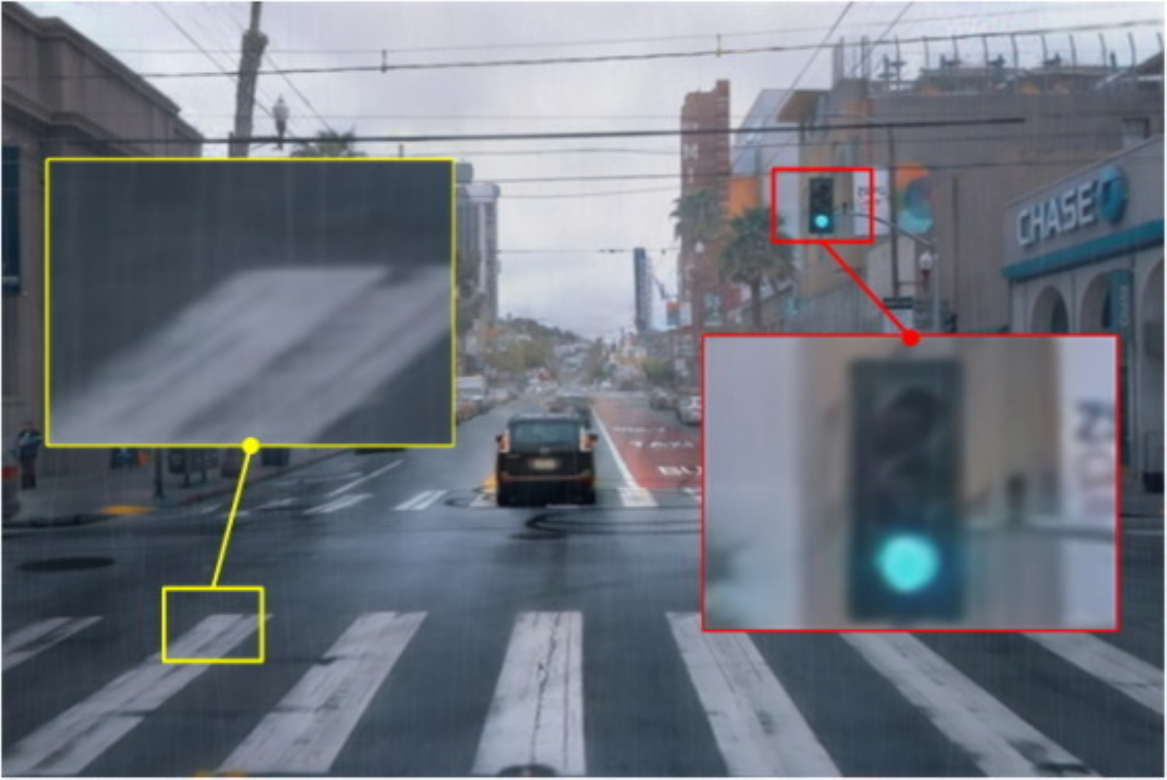}\\[-1pt]
\multicolumn{2}{c}{\footnotesize (c) $R=12.0~\mathrm{mm/h}$}
\end{tabular}
\caption{Rainfall synthesis under three target intensities in two scenes. Higher intensity produces denser near-field rain streaks and stronger far-field haze.}
\label{fig:intensity-comparison}
\end{figure*}

\subsection{Downstream-Task Application Validation}
\label{subsec:downstream}

The preceding experiments reconstruct 3D traffic scenes and synthesize rainy traffic conditions from natural-driving datasets. This section evaluates the practical use of the reconstructed scenes for autonomous-driving assessment through closed-loop end-to-end driving and open-loop object detection.

\subsubsection{Closed-Loop Simulation Framework}
\label{subsubsec:closed-loop-framework}

\begin{figure*}[!tb]
\centering
\includegraphics[width=0.80\textwidth]{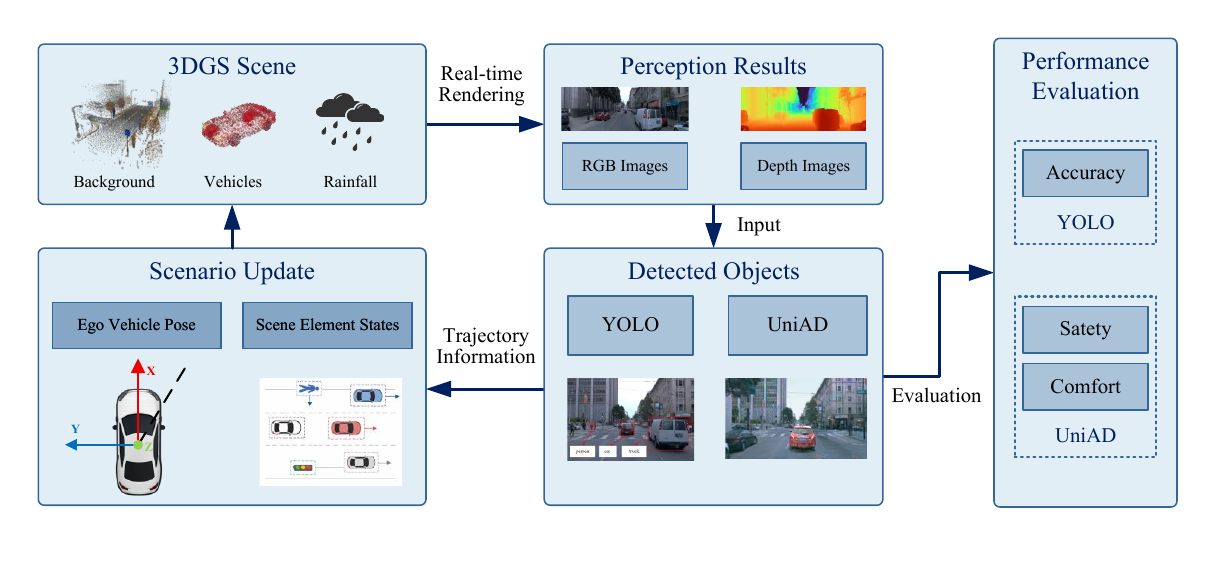}
\caption{3DGS-based closed-loop simulation and evaluation framework.}
\label{fig:closed-loop-framework}
\end{figure*}

To examine the applicability of 3DGS scenes to autonomous-driving system testing, we build a 3DGS-based closed-loop simulation framework on HUGSIM \cite{ref46}. The reconstructed traffic scene serves as the core of the framework, which integrates scene rendering, traffic-behavior simulation, weather-disturbance modeling, and the decision process of the autonomous-driving system into a unified simulation loop, as shown in Fig.~\ref{fig:closed-loop-framework}.

\subsubsection{Closed-Loop Evaluation of an End-to-End System}
\label{subsubsec:end-to-end-evaluation}

To further evaluate the testing capability of the constructed 3DGS simulation environment, two scenarios are considered: a cut-in leading vehicle and a stationary leading vehicle. The autonomous-driving system is evaluated under clear weather and three rainfall intensities.

\paragraph{Experimental Setup}
UniAD \cite{ref47} is used as the system under test (SUT). It is a planning-oriented end-to-end autonomous-driving framework that takes images as input and outputs planned trajectory points.

\textit{1) Evaluation Metrics:}

\emph{Safety.} The planned-point collision rate and time to collision (TTC) \cite{ref48} are used to evaluate SUT safety. The planned-point collision rate measures the safety of the future trajectory points generated by the SUT:
\begin{equation}
C=\frac{1}{FN}\sum_{i=1}^{F}\sum_{j=1}^{N}c_{ij}\times100\%,
\label{eq:collision-rate}
\end{equation}
\begin{equation}
c_{ij}=\begin{cases}
0, & \text{the $j$th planned point is collision-free},\\
1, & \text{the $j$th planned point causes a collision},
\end{cases}
\label{eq:collision-indicator}
\end{equation}
where $F$ is the number of frames and $N$ is the number of planned points in each frame. UniAD outputs five planned points per frame in the present experiments.

TTC is computed as
\begin{equation}
\mathrm{TTC}=\min\!\left(\frac{d_x}{v_{\mathrm{SUT}}-v_{\mathrm{BV}}},10\right),
\label{eq:ttc}
\end{equation}
where $v_{\mathrm{SUT}}$ and $v_{\mathrm{BV}}$ denote the speeds of the SUT and the background vehicle (BV), respectively, and $d_x$ is their longitudinal distance. When no collision risk exists, TTC is set to 10~s.

\emph{Comfort.} The comfort score measures whether the SUT motion parameters remain within predefined comfortable ranges:
\begin{equation}
S=\frac{1}{F}\sum_{i=1}^{F}s_i,
\label{eq:comfort-score}
\end{equation}
\begin{equation}
s_i=\begin{cases}
1, & \text{if all motion parameters satisfy their thresholds},\\
0, & \text{otherwise}.
\end{cases}
\label{eq:comfort-indicator}
\end{equation}
Following HUGSIM \cite{ref46}, the motion parameters and thresholds used for comfort evaluation are listed in Table~\ref{tab:comfort-thresholds}.

\begin{table}[!t]
\caption{Motion Parameters and Thresholds for Comfort Evaluation}
\label{tab:comfort-thresholds}
\centering
\footnotesize
\renewcommand{\arraystretch}{1.08}
\setlength{\tabcolsep}{3.2pt}
\begin{tabularx}{\columnwidth}{@{}>{\raggedright\arraybackslash}Xccc@{}}
\toprule
Parameter & Min. & Max. & Unit \\
\midrule
Longitudinal acceleration & $-4.05$ & $2.40$ & m/s$^2$ \\
Absolute longitudinal jerk & -- & $8.37$ & m/s$^3$ \\
Lateral acceleration & -- & $4.89$ & m/s$^2$ \\
Absolute yaw rate & -- & $0.95$ & rad/s \\
Absolute yaw acceleration & -- & $1.93$ & rad/s$^2$ \\
\bottomrule
\end{tabularx}
\end{table}

\begin{figure*}[!tb]
\centering
\subfloat[Cut-in leading-vehicle scenario]{\includegraphics[width=0.455\textwidth]{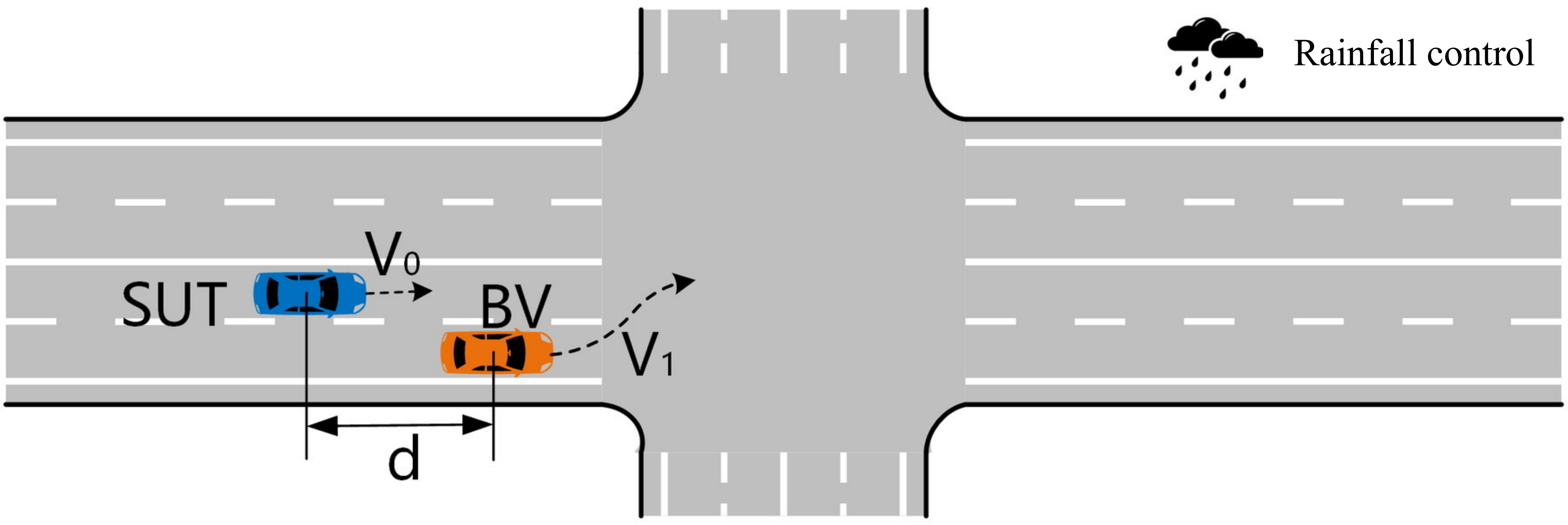}}
\hfil
\subfloat[Stationary leading-vehicle scenario]{\includegraphics[width=0.455\textwidth]{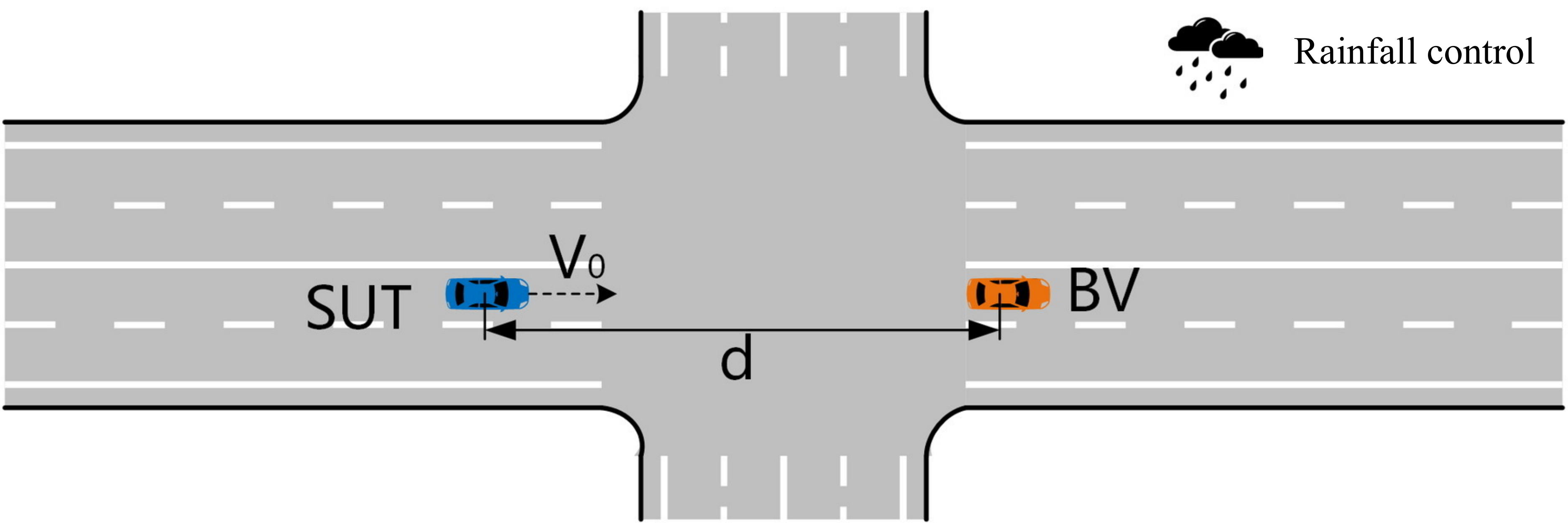}}
\caption{Closed-loop test scenarios in which the ego vehicle travels straight. $V_0$ and $V_1$ are the initial speeds of the SUT and BV, respectively, and $d$ is their initial longitudinal distance.}
\label{fig:test-scenarios}
\end{figure*}

\textit{2) Test Scenarios:}
The two test scenarios are shown in Fig.~\ref{fig:test-scenarios}. Each scenario controls the rainfall level by adjusting the numerical rainfall intensity.

The initial parameters of the cut-in and stationary-leading-vehicle scenarios are given in Tables~\ref{tab:cutin-settings} and~\ref{tab:stationary-settings}, respectively.

\begin{table}[!t]
\caption{Initial Parameters for the Cut-in Scenario}
\label{tab:cutin-settings}
\centering
\footnotesize
\renewcommand{\arraystretch}{1.08}
\setlength{\tabcolsep}{2.5pt}
\begin{tabularx}{\columnwidth}{@{}c*{4}{>{\centering\arraybackslash}X}@{}}
\toprule
No. & $V_0$ (m/s) & $d$ (m) & $V_1$ (m/s) & $R$ (mm/h) \\
\midrule
1 & 5 & 10 & 5 & 0 (clear) \\
2 & 5 & 10 & 5 & 1.2 \\
3 & 5 & 10 & 5 & 8.5 \\
4 & 5 & 10 & 5 & 12.0 \\
\bottomrule
\end{tabularx}
\end{table}

\begin{table}[!t]
\caption{Initial Parameters for the Stationary-Leading-Vehicle Scenario}
\label{tab:stationary-settings}
\centering
\footnotesize
\renewcommand{\arraystretch}{1.08}
\setlength{\tabcolsep}{3pt}
\begin{tabularx}{\columnwidth}{@{}c*{3}{>{\centering\arraybackslash}X}@{}}
\toprule
No. & $V_0$ (m/s) & $d$ (m) & $R$ (mm/h) \\
\midrule
5 & 10 & 40 & 0 (clear) \\
6 & 10 & 40 & 1.2 \\
7 & 10 & 40 & 8.5 \\
8 & 10 & 40 & 12.0 \\
\bottomrule
\end{tabularx}
\end{table}

\paragraph{Experimental Results}
The following experiments report UniAD performance in the cut-in and stationary-leading-vehicle scenarios and analyze its safety and comfort under different rainfall intensities.

\textit{1) Cut-in Scenario:}
The evaluation results are listed in Table~\ref{tab:cutin-results}. No collision occurs under clear weather or at 1.2~mm/h, although the minimum TTC values are small. Collisions occur at 8.5 and 12.0~mm/h.

\begin{table}[!t]
\caption{Evaluation Results for the Cut-in Scenario}
\label{tab:cutin-results}
\centering
\footnotesize
\renewcommand{\arraystretch}{1.08}
\setlength{\tabcolsep}{3.7pt}
\begin{tabular}{@{}cccccc@{}}
\toprule
Scene & $R$ & Coll. & Min. TTC & $C$ & $S$ \\
 & (mm/h) & & (s) & & \\
\midrule
1 & Clear & No & 0.110 & 8.9\% & 0.500 \\
2 & 1.2 & No & 0.151 & 8.9\% & 0.571 \\
3 & 8.5 & Yes & 0.000 & -- & -- \\
4 & 12.0 & Yes & 0.000 & -- & -- \\
\bottomrule
\end{tabular}
\end{table}

\begin{figure}[!t]
\centering
\includegraphics[width=0.98\linewidth]{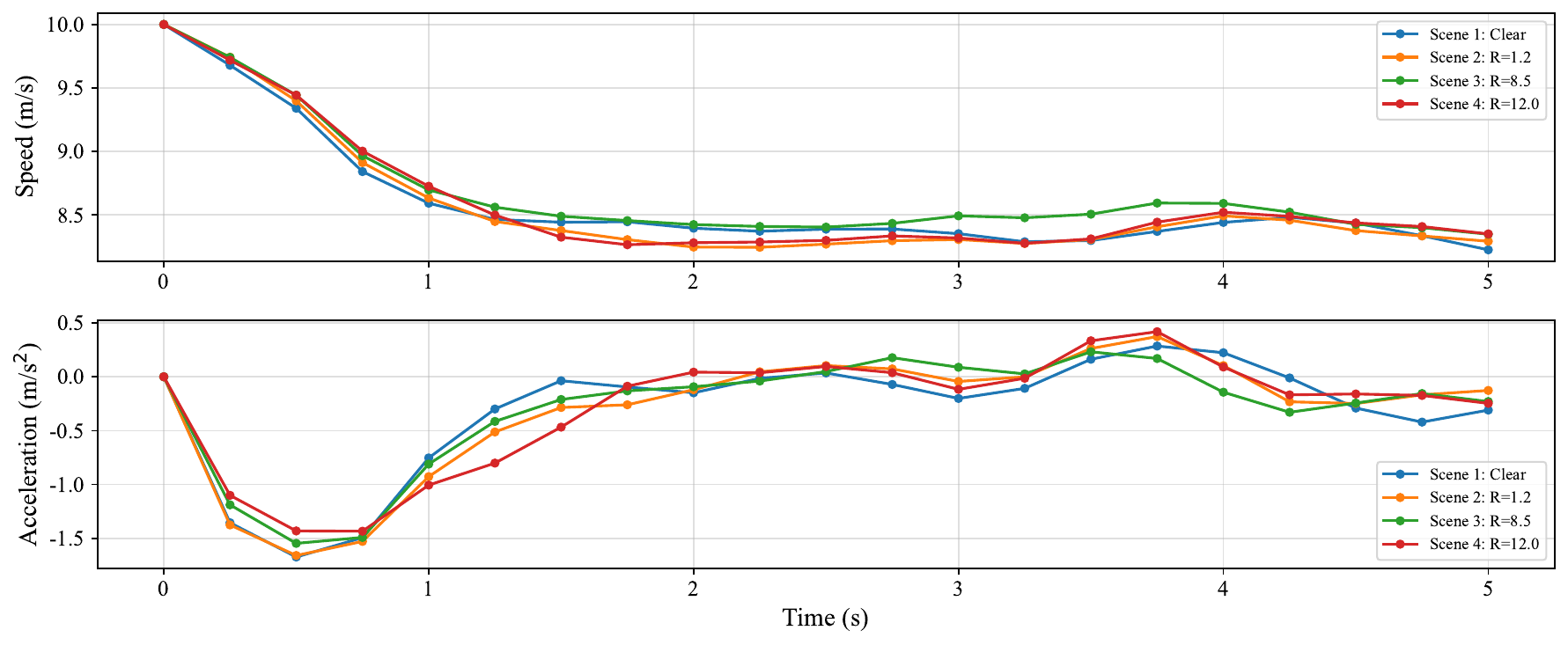}
\caption{SUT speed and acceleration in the cut-in scenario.}
\label{fig:cutin-kinematics}
\end{figure}

As shown in Fig.~\ref{fig:cutin-kinematics}, the absolute acceleration of the SUT is generally smaller in rainy scenes. During $t=1$--4~s, the SUT speed is generally higher than that under clear weather, eventually resulting in a collision with the BV. This result indicates that simulated rainfall reduces the risk-perception capability of the SUT to some extent.

\begin{figure}[!t]
\centering
\includegraphics[width=0.98\linewidth]{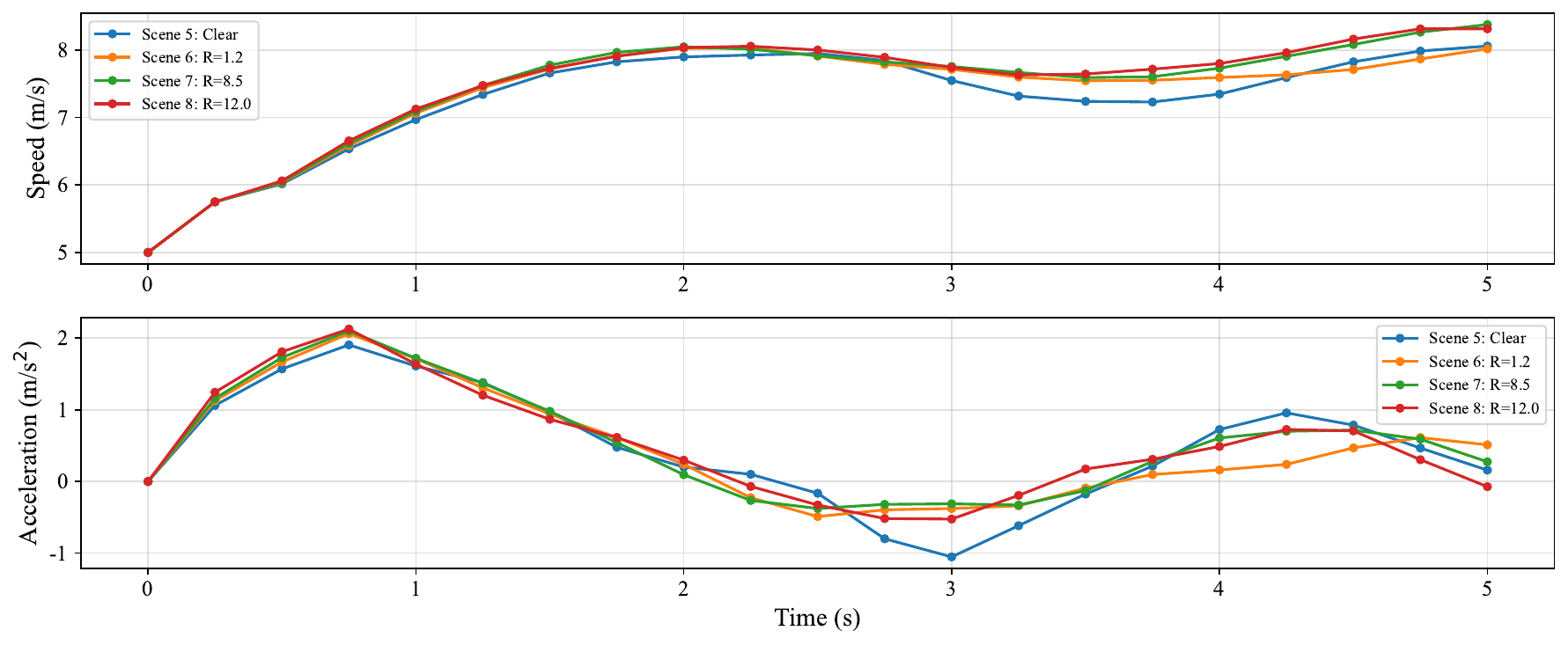}
\caption{SUT speed and acceleration in the stationary-leading-vehicle scenario.}
\label{fig:stationary-kinematics}
\end{figure}

\textit{2) Stationary-Leading-Vehicle Scenario:}
Table~\ref{tab:stationary-results} shows that no collision occurs at any tested rainfall intensity. Across the rainy conditions, however, the lowest minimum TTC, highest planned-point collision rate, and lowest comfort score occur at different rainfall intensities, indicating that both SUT safety and comfort may deteriorate in rainy weather. Figure~\ref{fig:stationary-kinematics} further shows that acceleration and deceleration responses are delayed relative to clear weather, increasing driving risk.

\begin{table}[!t]
\caption{Evaluation Results for the Stationary-Leading-Vehicle Scenario}
\label{tab:stationary-results}
\centering
\footnotesize
\renewcommand{\arraystretch}{1.08}
\setlength{\tabcolsep}{3.7pt}
\begin{tabular}{@{}cccccc@{}}
\toprule
Scene & $R$ & Coll. & Min. TTC & $C$ & $S$ \\
 & (mm/h) & & (s) & & \\
\midrule
5 & Clear & No & 0.246 & 7.1\% & 0.881 \\
6 & 1.2 & No & 0.263 & 7.5\% & 0.875 \\
7 & 8.5 & No & 0.175 & 4.9\% & 0.854 \\
8 & 12.0 & No & 0.253 & 7.3\% & 0.805 \\
\bottomrule
\end{tabular}
\end{table}

\subsubsection{Open-Loop Evaluation of Object-Detection Algorithms}
\label{subsubsec:object-detection}

\paragraph{Experimental Setup}
To evaluate how the reconstructed 3DGS dynamic traffic scenes and simulated rainfall scenes affect object detection, YOLOv10n \cite{ref49} and YOLO26n \cite{ref50} are tested in the original scenes, reconstructed 3DGS scenes, and simulated rainy scenes at different rainfall intensities. Images from two urban intersections are input to the detectors, which output bounding boxes, class labels, and confidence scores for evaluating object detection and recognition.

\begin{figure}[!t]
\centering
\subfloat[Intersection 1]{\includegraphics[width=0.49\linewidth]{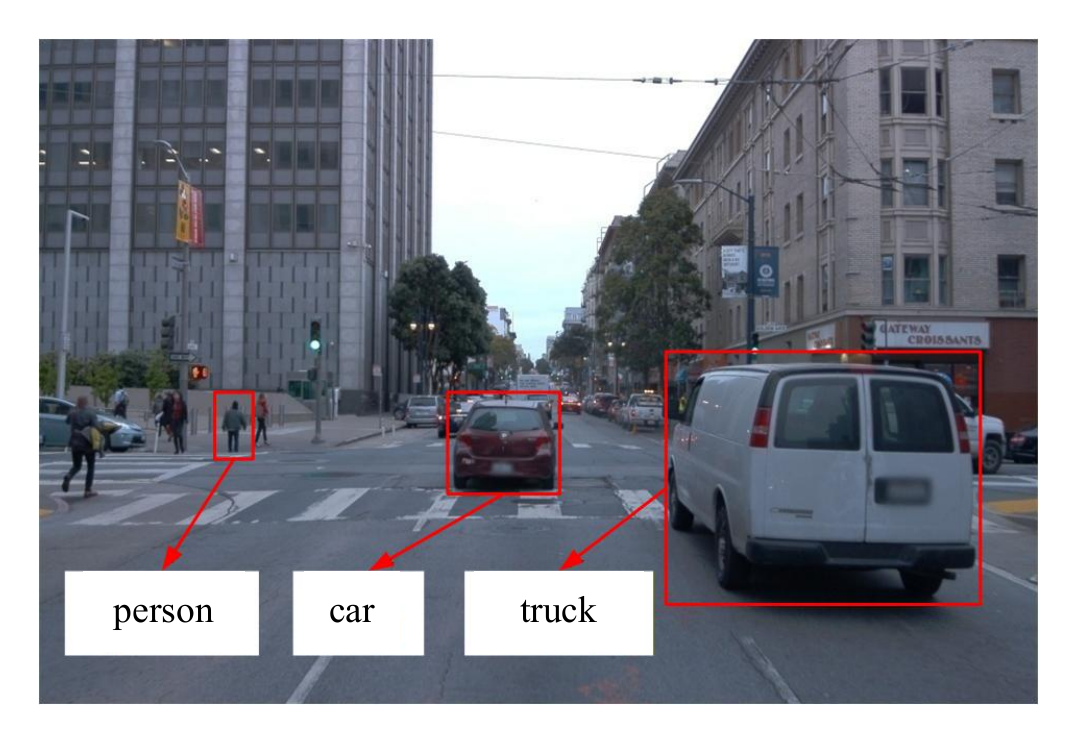}}\hfil
\subfloat[Intersection 2]{\includegraphics[width=0.49\linewidth]{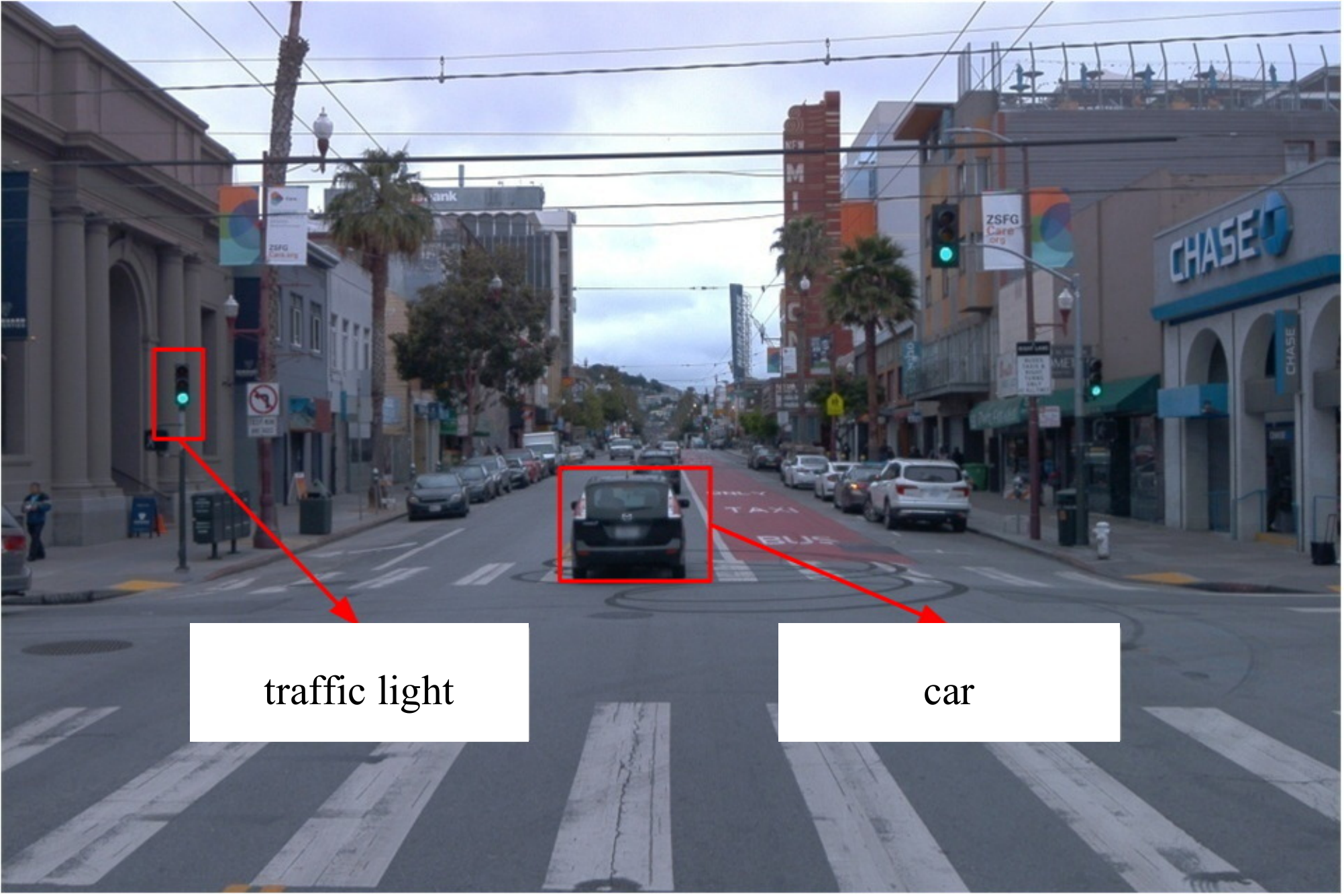}}
\caption{Ground-truth annotations of the evaluated objects at two urban intersections.}
\label{fig:ground-truth}
\end{figure}

\paragraph{Experimental Results}
The two models are compared at two urban intersections, with detailed analyses of car and truck predictions. The ground-truth labels are shown in Fig.~\ref{fig:ground-truth}.

\begin{figure}[!t]
\centering
\subfloat[Intersection 1]{\includegraphics[width=\linewidth]{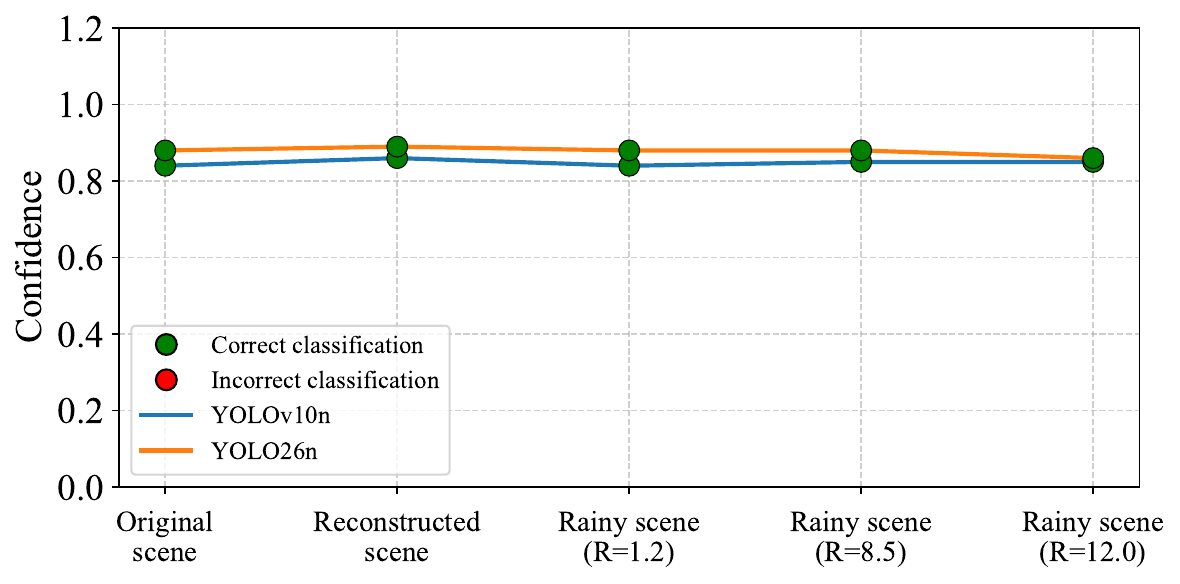}}\\[2pt]
\subfloat[Intersection 2]{\includegraphics[width=\linewidth]{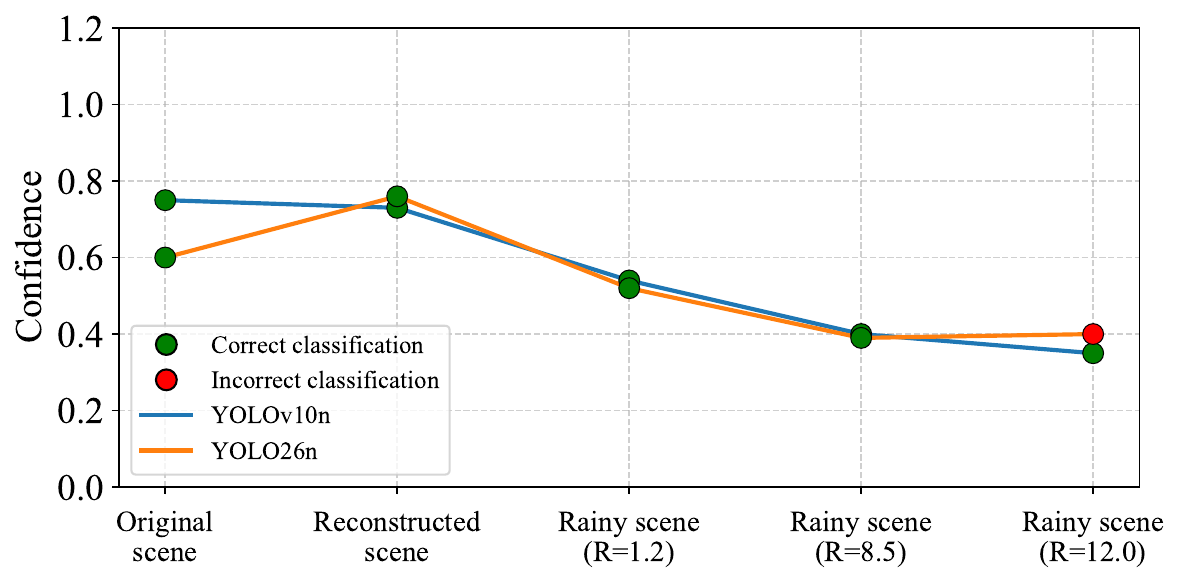}}
\caption{Car-classification confidence under the original, reconstructed, and simulated-rainfall conditions.}
\label{fig:car-confidence}
\end{figure}

\textit{1) Car Detection Results:}
Figure~\ref{fig:car-confidence} compares car-classification confidence under the original, reconstructed, and simulated-rainfall conditions.

At Intersection~1, the car is close to the camera and light in color. The classification confidence of both YOLOv10n and YOLO26n therefore remains stable at approximately 0.85, with only small fluctuations. At Intersection~2, the car is farther away and darker. As rainfall intensity increases, the confidence of both models decreases from approximately 0.6--0.8 to about 0.4. At 12.0~mm/h, YOLO26n misclassifies the target as a truck.

These results show that the generated rainfall effects can degrade object-detection performance and that the confidence reduction with increasing rainfall intensity follows the expected tendency in real rainy environments. For vehicle targets, the prediction differences between the reconstructed and original scenes are small, further supporting the realism of the reconstructed results in both visual appearance and task-level influence.

\begin{figure}[!t]
\centering
\includegraphics[width=\linewidth]{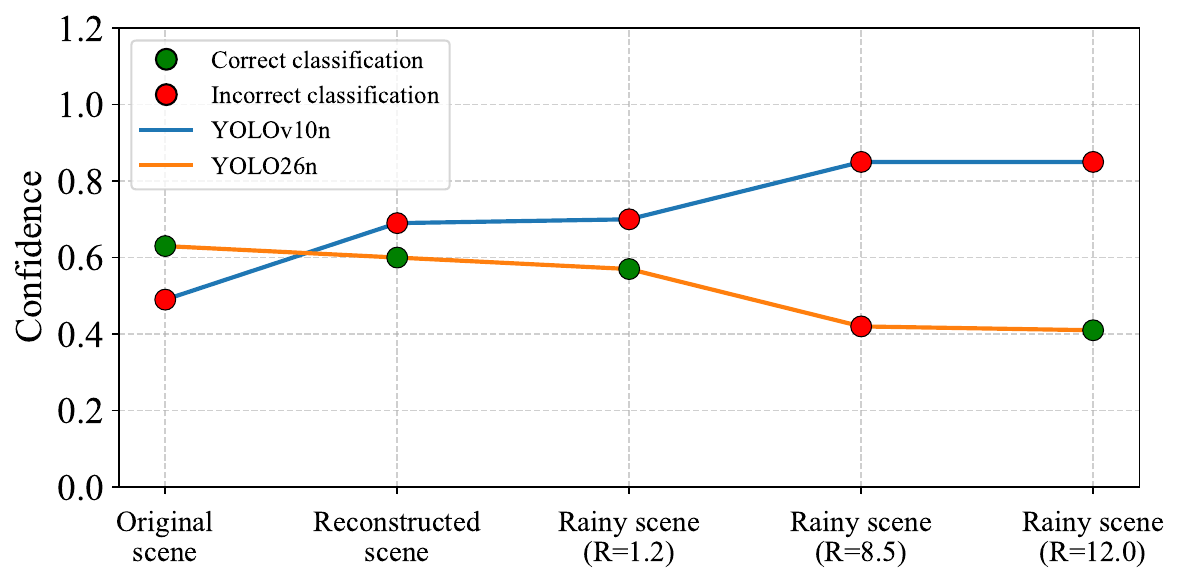}
\caption{Truck-classification confidence at Intersection~1.}
\label{fig:truck-confidence}
\end{figure}

\textit{2) Truck Detection Results:}
Figure~\ref{fig:truck-confidence} shows the truck-classification confidence at Intersection~1.

Although the truck is close to the camera, YOLOv10n fails to recognize it correctly, and its confidence in the incorrect car class increases with rainfall intensity. YOLO26n correctly recognizes the truck in the original scene, but its confidence decreases from 0.63 to 0.41 as rainfall intensity increases, and the model misclassifies the truck as a car at 8.5~mm/h.

\FloatBarrier
\section{Conclusion}
\label{sec:conclusion}

This paper presents GSRAIN, a physically calibrated high-/low-frequency rainfall synthesis method for 3DGS driving scenes. Measured rainfall data establish explicit mappings from raindrop diameter, velocity, and number density to the target rainfall intensity. Geometry-constrained, cross-view-consistent rainy-appearance transfer provides the low-frequency component, while global rainy appearance, near-field rain streaks, and far-field haze are jointly represented in a unified 3DGS scene with rainfall-intensity control over 0--13~mm/h. GSRAIN achieves an FID of 149.09, lower than 155.71 for CycleGAN-Turbo and 157.94 for WeatherEdit. Object-detection and closed-loop driving experiments further show differentiated responses in perception predictions and driving behavior; in particular, collisions occur in the cut-in scenario at 8.5 and 12.0~mm/h. These results indicate that GSRAIN provides physically calibrated, repeatable, and multi-view-renderable 3D rainy scenes for rainfall-sensitivity analysis of autonomous-driving algorithms. Future work will investigate direct 3D modeling of low-frequency rainy appearance and consistency optimization for cross-domain 3DGS assets.

\FloatBarrier
\IEEEtriggeratref{46}

\end{document}